\documentclass{article} 
\usepackage[final]{colm2026_conference}

\usepackage[utf8]{inputenc} 
\usepackage{longtable}
\usepackage[T1]{fontenc}    
\usepackage{hyperref}       
\usepackage{url}            
\usepackage{booktabs}       
\usepackage{amsfonts}       
\usepackage{nicefrac}       
\usepackage{microtype}      
\usepackage[dvipsnames]{xcolor}
\usepackage[most]{tcolorbox}
\usepackage{colortbl}
\usepackage{graphicx}
\usepackage{natbib}
\usepackage{adjustbox}
\usepackage{multicol}
\usepackage{multirow}
\usepackage{arydshln}
\usepackage{bbding}
\usepackage{pifont}
\usepackage{listings}
\usepackage{caption}
\usepackage{enumitem}
\usepackage{amsthm}

\usepackage{lineno}

\definecolor{darkblue}{rgb}{0, 0, 0.5}
\hypersetup{colorlinks=true, citecolor=darkblue, linkcolor=darkblue, urlcolor=darkblue}

\definecolor{lgra}{HTML}{F0F0EB}
\definecolor{lora}{HTML}{FFD2A4}
\definecolor{lgre}{HTML}{A4FFAE}
\definecolor{sgre}{HTML}{3EFF54}
\definecolor{lred}{HTML}{FFA4A4}
\definecolor{sred}{HTML}{FF7171}
\definecolor{lblu}{HTML}{A4C7FF}
\definecolor{mplred}{HTML}{D62728}

\definecolor{tcolorBlue}{RGB}{203, 231, 254} 
\definecolor{bleudefrance}{rgb}{0.19, 0.55, 0.91}
\definecolor{mediumorchid}{rgb}{0.73, 0.33, 0.83}
\definecolor{electriclavender}{rgb}{0.96, 0.73, 1.0}
\definecolor{babyblue}{rgb}{0.54, 0.81, 0.94}
\definecolor{lightcyan}{rgb}{0.88, 1.0, 1.0}

\newcommand{\ie}{\textit{i.e., }}
\newcommand{\eg}{\textit{e.g., }}

\newcommand{\cmark}{\textcolor{OliveGreen}{\Checkmark}}
\newcommand{\xmark}{\textcolor{red}{\ding{56}}}
\newcommand{\trendcell}[2]{\cellcolor{#1}#2}
\newcommand{\inlinecolor}[2]{\colorbox{#1}{\raisebox{0pt}[0.8ex][0ex]{#2}}}

\newtheorem{theorem}{Theorem}

\newtheorem{corollary}{Corollary}

\theoremstyle{definition}
\newtheorem{definition}{Definition}

\theoremstyle{remark}
\newtheorem{remark}{Remark}

\title{Test-Time Scaling in Reasoning Models Is Not \\Effective for Knowledge-Intensive Tasks Yet}

\author{James Xu Zhao \qquad Bryan Hooi \qquad See-Kiong Ng \\
National University of Singapore\\
\texttt{xu.zhao@u.nus.edu} \qquad \texttt{\{dcsbhk, seekiong\}@nus.edu.sg} 
}

\begin{document}

\ifcolmsubmission
\linenumbers
\fi

\maketitle

\begin{abstract}
Test-time scaling increases inference-time computation through longer reasoning chains and has shown strong performance gains across many domains. However, frontier models still suffer from factuality hallucinations, raising the question of whether increased computation is effective on closed-book knowledge-intensive tasks. In this work, we evaluate 14 reasoning models under different test-time scaling strategies. Our results challenge its effectiveness: increasing test-time computation does not consistently improve accuracy and often leads to more hallucinations. We find that changes in hallucination rates are largely driven by the model's willingness to answer, as longer reasoning encourages more attempts, many of which are incorrect. We also observe patterns consistent with confirmation bias, where extended reasoning reinforces early incorrect beliefs with fabricated details. Finally, we provide an information-theoretic perspective showing that compute-only test-time scaling, as a post-processing procedure of a fixed model, cannot introduce new information about the ground-truth answer, explaining the limited performance gains. Overall, our findings highlight important limitations of current test-time scaling methods for closed-book knowledge-intensive tasks.\footnote{Our code and data are available at: \url{https://github.com/XuZhao0/tts-knowledge}.}
\end{abstract}

\section{Introduction}
Reasoning models, such as GPT-5, Gemini 2.5, and Qwen3, have demonstrated impressive performance on many challenging benchmarks, ranging from competition mathematics to complex programming tasks~\citep{rein2024gpqa, petrov2025proof, phan2025lastexam, jain2025livecodebench, zaremba2025trading}. A key technique behind these breakthroughs is test-time scaling, where the model is allocated more inference-time compute to perform extensive reasoning, such as generating longer chain-of-thought (CoT) reasoning traces~\citep{o1blog, comanici2025gemini, balachandran2025inferencetimescalingcomplextasks}.

Despite these advances, frontier models still suffer from hallucinations, generating statements that contradict world knowledge~\citep{wang-etal-2024-factuality, Augenstein2024FactualityCI, 10.1145/3703155}. This remains a fundamental challenge because many practical applications, such as healthcare, rely on models to provide accurate information, and hallucinations in these settings can directly undermine reliability and create high-stakes risks~\citep{10.1093/jla/laae003, kim2025medicalhallucinationsfoundationmodels}. To study this issue, knowledge-intensive tasks are widely used to evaluate models' ability to recall factual information from internal knowledge. Given that test-time scaling has shown promise across many domains, a natural question arises: 

\begin{center}
\textbf{\textit{Is test-time scaling effective for closed-book knowledge-intensive tasks?}}
\end{center}

To answer this question, we conduct a comprehensive study of test-time scaling on knowledge-intensive tasks that require models to rely solely on their parametric knowledge, without access to external evidence or tools. We evaluate 14 reasoning models under different test-time scaling strategies. Contrary to expectations, our results, summarized in Table~\ref{tab:summary_test_time}, show that increasing test-time computation does not consistently improve accuracy or reduce hallucinations for most models.\footnote{Enabling thinking can still be beneficial compared to non-thinking mode. It is the marginal return from additional test-time reasoning that flattens. See Appendix~\ref{sec:think_vs_no_think}.} In some cases, such as GPT-5 mini, increasing thinking time even increases hallucinations. Our analysis suggests that these increases are largely driven by the model attempting more questions under extended reasoning. We further observe that longer reasoning can induce confirmation bias, where models generate fabricated details to support prior beliefs, resulting in overconfident hallucinations.

To explain the empirical results, we provide an information-theoretic perspective. We model these tasks as \textit{arbitrary facts} questions, where correct answers cannot be inferred from the prompt structure. Under this formulation, compute-only test-time scaling is a post-processing procedure applied to a fixed model, so it cannot introduce new information about the ground-truth answer beyond what the model already contains. This yields an information-theoretic upper bound on performance in information-limited regimes, explaining the limited accuracy gains from additional test-time compute. Together with the empirical observation that longer reasoning can increase the answer rate, this bound also helps explain why hallucinations may rise when the error rate among attempted answers does not decrease sufficiently.

In summary, while test-time scaling has led to impressive performance gains across many domains, our work shows that it is \textit{not} yet effective for knowledge-intensive tasks from both empirical and theoretical perspectives, suggesting important limitations in this setting.

\begin{table}
\centering
\resizebox{0.92\textwidth}{!}{
\begin{tabular}{l*{7}{>{\centering\arraybackslash}p{1.7cm}}}
\toprule
\textbf{Metric} & \textbf{GPT-5} & \textbf{GPT-5 mini} & \textbf{o3-mini} & \textbf{o4-mini} & \textbf{gpt-oss-20b} & \textbf{Grok-3 mini} & \textbf{Gemini 2.5 Flash}   \\
\midrule
\textsc{Accuracy}  & \trendcell{lora}{$\sim$}        & \trendcell{lora}{$\sim$}          & \trendcell{lora}{$\sim$}      & \trendcell{lora}{$\sim$}       &  \trendcell{lora}{$\sim$}      & \trendcell{lora}{$\sim$}     & \trendcell{lgre}{↑}        \\
\textsc{Hallucination}    & \trendcell{lora}{$\sim$}  & \trendcell{lred}{↑}           & \trendcell{lred}{↑}      & \trendcell{lora}{$\sim$}        & \trendcell{lred}{↑}      & \trendcell{lgre}{↓}      & \trendcell{lred}{↑}     \\
\midrule
\midrule
\textbf{Metric} & \textbf{Gemini 2.5 Pro} & \textbf{Claude Sonnet 4} & \textbf{R1-Distill- Qwen-7B} & \textbf{R1-Distill- Qwen-14B} & \textbf{R1-Distill- Llama-8B} & \textbf{Qwen3-8B} & \textbf{Qwen3-14B}  \\
\midrule
\textsc{Accuracy}   & \trendcell{lgre}{↑}      & \trendcell{lora}{$\sim$}          & \trendcell{lora}{$\sim$}      & \trendcell{lora}{$\sim$}       &  \trendcell{lora}{$\sim$}      & \trendcell{lora}{$\sim$}     & \trendcell{lora}{$\sim$}         \\
\textsc{Hallucination}   & \trendcell{lora}{$\sim$}  & \trendcell{lora}{$\sim$}           & \trendcell{lora}{$\sim$}     & \trendcell{lgre}{↓}       & \trendcell{lora}{$\sim$}   & \trendcell{lora}{$\sim$}  & \trendcell{lora}{$\sim$}   \\
\bottomrule
\end{tabular}
}
\caption{Summary of model behavior with test-time scaling. Increasing test-time computation does not consistently improve accuracy or reduce hallucinations for most models. For \textsc{accuracy}, \inlinecolor{lgre}{$\uparrow$} denotes consistent improvement with >2\% accuracy gains across consecutive reasoning levels, while \inlinecolor{lora}{$\sim$} indicates no consistent trend. For \textsc{hallucination}, \inlinecolor{lgre}{$\downarrow$} denotes consistent reduction with >2\% hallucination decrease, \inlinecolor{lred}{$\uparrow$} indicates degradation with >2\% hallucination increase, and \inlinecolor{lora}{$\sim$} reflects inconsistent or fluctuating patterns.}
\label{tab:summary_test_time}
\end{table}

\section{Related Work}
\subsection{Test-Time Scaling} 
Test-time scaling has emerged as a promising strategy for enhancing the capabilities of large language models. It is typically categorized into two main paradigms~\citep{zhang2025surveytesttimescalinglarge}: (1) the parallel approach, which samples multiple outputs independently and aggregates them~\citep{brown2024largelanguagemonkeysscaling, snell2025scaling}; and (2) the sequential approach, where the model generates long chain-of-thought (CoT) reasoning traces before producing an answer~\citep{NEURIPS2022_9d560961, o1blog, muennighoff2025s10, yang2025demystifying}. In this work, we focus on the sequential paradigm, which has become the dominant test-time scaling method for improving model performance.~\footnote{Results and analysis for the parallel approach are provided in Appendix~\ref{app:parallel}.} It is widely adopted in frontier reasoning models~\citep{gpt-5, comanici2025gemini, claude4, grok3blog, guo2025deepseek, yang2025qwen3}, and has demonstrated strong performance across a range of challenging tasks, such as competition mathematics~\citep{aime_1983_2024,petrov2025proof,jain2025livecodebench,rein2024gpqa}. 

However, recent studies suggest that increasing test-time computation does not always improve performance in some tasks~\citep{fu2025scaling, li2025thinking, liu2025thinkingseeingassessingamplified}. \citet{gema2025inversescalingtesttimecompute} identify inverse scaling, where longer reasoning may reinforce problematic patterns rather than improve accuracy. \citet{cuadron2025danger} reveal that excessive internal reasoning reduces effectiveness in agentic tasks. \citet{wu2026when} find that reasoning performance follows an inverted-U relationship with CoT length, where longer reasoning can degrade accuracy due to factors such as error accumulation. Our study aligns with this line of work, showing ``negative'' scaling of test-time computation in knowledge-intensive tasks.

\subsection{Factuality Hallucinations in LLMs}
Factuality hallucinations, which refer to content that contradicts world knowledge, have been a long-standing issue in large language models~\citep{10.1145/3571730, zhang2023siren0s, 10.1145/3703155, openai_safety_tests}. These hallucinations reflect the models' limitations in absorbing knowledge and their inability to recognize knowledge boundaries~\citep{bang-etal-2025-hallulens}. In high-stakes domains, such as healthcare and law, hallucinations can introduce serious risks~\citep{10.1093/jla/laae003, kim2025medicalhallucinationsfoundationmodels}.
Prior studies reveal that longer responses often lead to lower factual precision~\citep{NEURIPS2024_937ae0e8, zhao-etal-2025-response}, while our work focuses on the effect of reasoning length on hallucinations. \citet{kalai2025languagemodelshallucinate} argue that LLMs hallucinate because current training and evaluation paradigms favor guessing over acknowledging uncertainty. We provide empirical evidence for this in test-time scaling settings and further provide an information-theoretic analysis that formalizes an information ceiling for test-time scaling. To the best of our knowledge, this is the first study to examine how increased test-time computation affects factuality hallucinations.

\section{How Does Test-Time Scaling Affect Accuracy and Hallucination Ratio?}
\label{sec:main_results_tts}
To understand the impact of test-time scaling on knowledge-intensive tasks, we evaluate how increasing test-time computation affects accuracy and hallucination ratio across 14 reasoning models on three benchmarks.

\subsection{Experimental Setup}
\label{sec:exp_setup_tts}

\paragraph{Benchmarks.} We evaluate three challenging knowledge-intensive benchmarks that cover diverse topics and measure whether models can accurately access their internal knowledge: (1) \textbf{SimpleQA}~\citep{wei2024measuring}, covering topics from science to arts; (2) \textbf{FACTS Parametric}~\citep{cheng2025facts}, containing questions reflecting real-world interests; (3) \textbf{FRAMES}~\citep{krishna-etal-2025-fact}, requiring internal knowledge and multi-hop reasoning.

\paragraph{Models and Test-Time Scaling Settings.} We evaluate 14 large reasoning models and group them into three categories, based on how they support test-time scaling. Models are not allowed to access any external knowledge sources.

\begin{itemize}[leftmargin=*, nolistsep]
\setlength{\itemsep}{1mm}
    \item \textbf{Reasoning effort}: Models such as GPT-5~\citep{gpt-5}, o3-mini, o4-mini~\citep{o3o4}, gpt-oss-20b~\citep{gpt-oss}, Grok-3 mini~\citep{grok3blog} accept a \texttt{reasoning effort} parameter that adjusts the time of thinking.
    \item \textbf{Thinking budget}: Models like Gemini 2.5~\citep{comanici2025gemini}, Claude Sonnet 4~\citep{claude4} provide a \texttt{thinking budget} parameter that guides the model on the number of thinking tokens to use. A larger number encourages the model to think for a longer time.
    \item \textbf{Budget forcing}: For DeepSeek-R1-Distill models~\citep{guo2025deepseek} and Qwen3 models~\citep{yang2025qwen3}, we adopt the budget forcing method~\citep{muennighoff2025s10}, which extends the model's thinking process by appending \texttt{``Wait''} when the model attempts to terminate its reasoning.
\end{itemize}

\paragraph{Prompts.} We use a consistent prompting format for all models, except GPT-5 mini and Claude Sonnet 4, across all benchmarks: \textit{Give me the answer to the following question only when you are sure of it. Otherwise, say `I don't know'. Put your answer on its own line after `Answer:'.}~\footnote{For GPT-5 mini and Claude Sonnet 4, this prompt leads to over 80\% abstention. Therefore, we do not explicitly instruct them to abstain when uncertain. See Appendix~\ref{appendix:prompt} for details. Results with an alternative prompt that does not instruct abstention are in Appendix~\ref{apd:other_prompt}.}

\paragraph{Evaluation and Metrics.} Following~\citet{wei2024measuring}, we use \texttt{gpt-4o-mini} as a grader to evaluate responses. For each question, the grader is provided with both the model's predicted answer and the reference answer, and assigns one of three labels: \textit{``correct''}, \textit{``incorrect''}, or \textit{``not attempted''}.~\footnote{We randomly sampled 300 cases and compared the automatic evaluation results with human annotations. We found only 2 disagreements.} We report two metrics: (1) \textit{Accuracy}, the percentage of all questions that were answered correctly. (2) \textit{Hallucination ratio}, the percentage of all questions that were answered incorrectly. The remaining questions are abstentions.  We additionally report F-score, which combines accuracy and hallucinations, in Appendix~\ref{ap:results_f_score}. More experimental details are provided in Appendix~\ref{appendix:implementation}.

\begin{figure}[t]
    \centering
    \includegraphics[width=\textwidth]{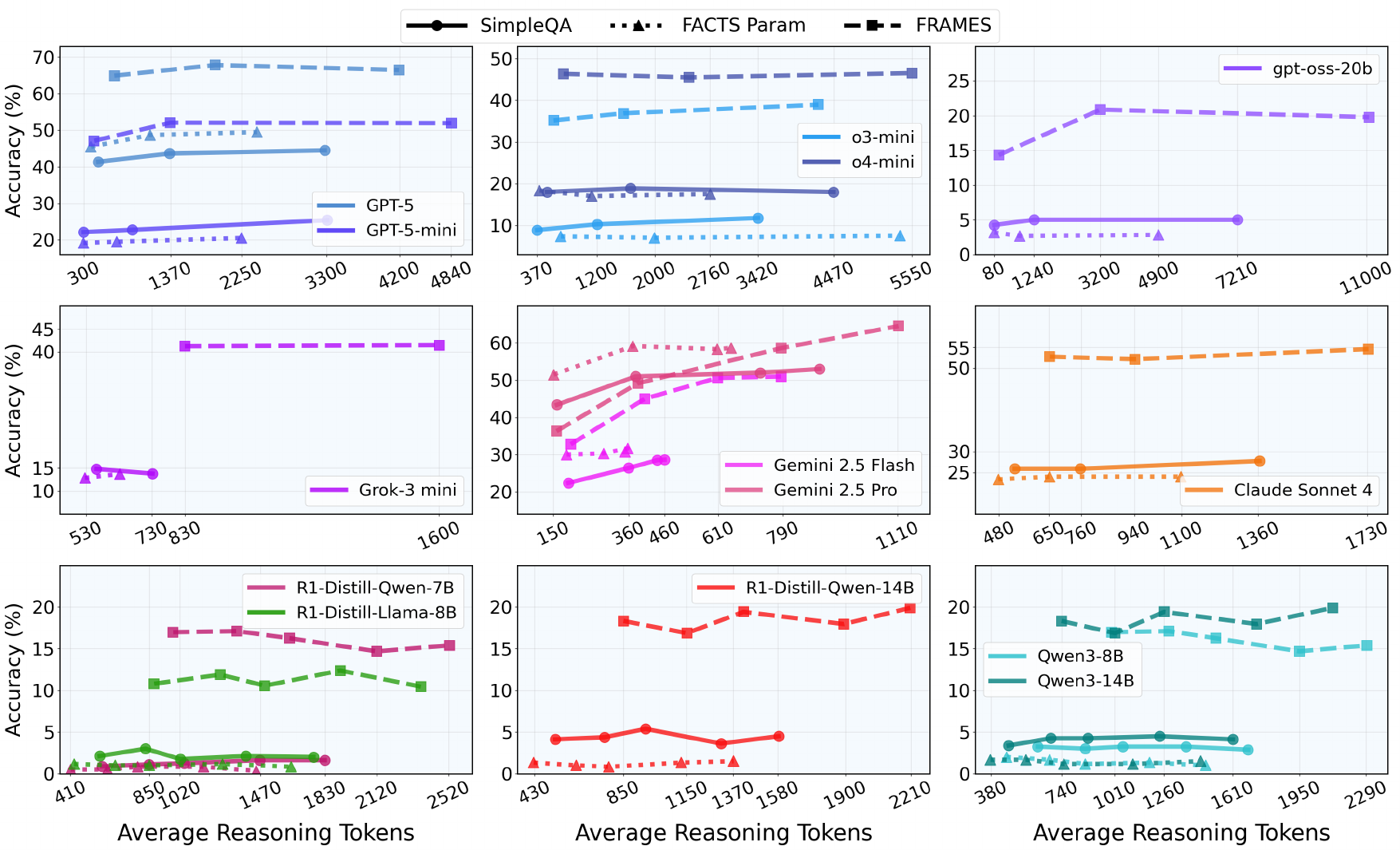}
    \caption{
    Accuracy with increased test-time computation across 14 reasoning models. For most models, extended test-time reasoning does not consistently improve accuracy. While some models, such as GPT-5, show initial accuracy gains, further increasing reasoning length brings little or no additional improvement.
    }
    \label{fig:accuracy_main}
\end{figure}

\subsection{Effect of Test-Time Scaling on Accuracy}
\label{sec:main_accuracy_results}

\paragraph{Thinking more does not consistently improve the accuracy for most models.} As shown in Figure~\ref{fig:accuracy_main}, increasing the reasoning length results in minimal or no accuracy gains across most models and all benchmarks. Higher \texttt{reasoning effort} leads to much longer thinking length, but does not consistently improve accuracy. For example, o4-mini exhibits more than 8 times increase in reasoning tokens, but the accuracy remains almost unchanged on all tasks. On FRAMES, GPT-5 and gpt-oss-20b show a 3-5\% accuracy increase when increasing effort from \textit{`low'} to \textit{`medium'}, but no further improvement at \textit{`high'} effort, despite the average reasoning tokens more than doubling. Increasing the \texttt{thinking budget} of Claude Sonnet 4 results in minimal accuracy gains, less than 2\% improvement, even as the average thinking length nearly triples. Among models using \texttt{budget forcing}, such as R1-Distill-Qwen-7B and Qwen3-14B, accuracy fluctuates as the thinking length increases. 

\paragraph{For Gemini 2.5, low thinking budget limits accuracy due to incomplete reasoning.} We evaluate Gemini 2.5 Flash and 2.5 Pro with \texttt{thinking budget} of 256, 512, 1024 and 2048 tokens. On SimpleQA, accuracy improves by over 6\% for both models, as the budget increases from 256 to 1024 tokens. However, increasing the budget further to 2048 tokens does not bring more improvement. On FRAMES, Gemini 2.5 achieves more accuracy gains. These improvements arise because Gemini 2.5 is often unable to complete its reasoning under low thinking budgets, which limits its accuracy. This issue is particularly evident on FRAMES, which requires more reasoning steps. We provide more analysis in Appendix~\ref{sec:gemini_case_study}.

\subsection{Effect of Test-Time Scaling on Hallucination Ratio}
\label{sec:hallucination_results}

\begin{figure}[t]
    \centering
    \includegraphics[width=\textwidth]{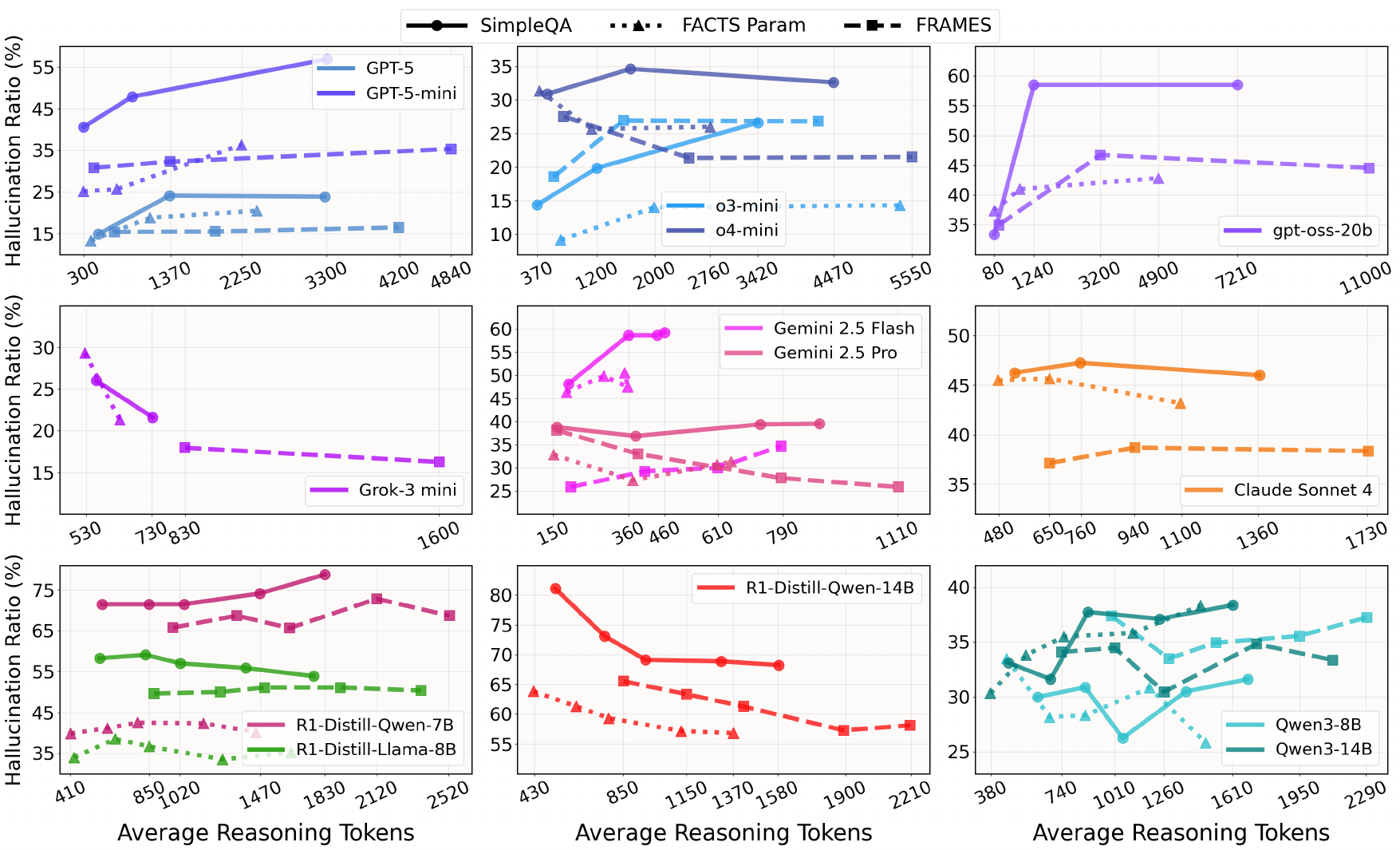}
    \caption{
    Hallucination ratio with increased test-time reasoning across 14 models. For most models, longer reasoning does not reduce hallucinations. In many cases, such as GPT-5 mini, hallucination increases with longer thinking length.
    }
    \label{fig:hallucination_main_results}
\end{figure}

\paragraph{Thinking more does not reduce hallucinations and may even increase them.} In Figure~\ref{fig:hallucination_main_results}, for most models, increasing test-time computation fails to reduce hallucinations and can make them worse. In OpenAI models, higher \texttt{reasoning effort} often leads to more hallucinations. For example, on SimpleQA, the hallucination ratio of GPT-5-mini increases by over 15\% as reasoning length increases from 300 to 3300 tokens. Similarly, GPT-5, o3-mini, and gpt-oss-20b show increases of 9\%, 12\%, and 25\%, respectively, as reasoning length scales by over 8 times. The same trend holds on the other tasks. Models with \texttt{thinking budget} settings exhibit similar behavior. For Gemini 2.5 Flash, hallucination ratio increases by over 5\% on all benchmarks. Claude Sonnet 4 shows minimal or no reduction in hallucinations despite longer thinking. Among models using \texttt{budget forcing}, none except for DS-R1-Distill-Qwen-14B benefits from longer thinking in terms of reducing hallucinations. 

\paragraph{For Grok-3 mini and R1-Distill-Qwen-14B, thinking more leads to fewer hallucinations.} Increasing the \texttt{reasoning effort} reduces the hallucination ratio of Grok-3 mini by 4.4\%, 8.0\% and 1.7\% on SimpleQA, FACTS Param and FRAMES, respectively. However, considering that the reasoning length nearly doubles on FRAMES, the reduction is relatively small. R1-Distill-Qwen-14B shows a more noticeable improvement. On SimpleQA, hallucinations drop by over 12\% as the average reasoning length increases from 530 to 1580 tokens. In the other two tasks, the hallucination ratio decreases by around 8\% as reasoning length increases. Despite these reductions, R1-Distill-Qwen-14B still has a higher hallucination ratio than other open-source models, such as R1-Distill-Llama-8B and Qwen3-8B.

\section{Why Does Thinking More Sometimes Lead to Fewer or More Hallucinations?}
\label{sec:causes_fewer_more_hallu}

As discussed in Section~\ref{sec:hallucination_results}, thinking more sometimes leads to fewer hallucinations, and sometimes more. In this section, we examine the causes of these changes. 

\begin{figure}[t]
    \centering
    \includegraphics[width=0.95\textwidth]{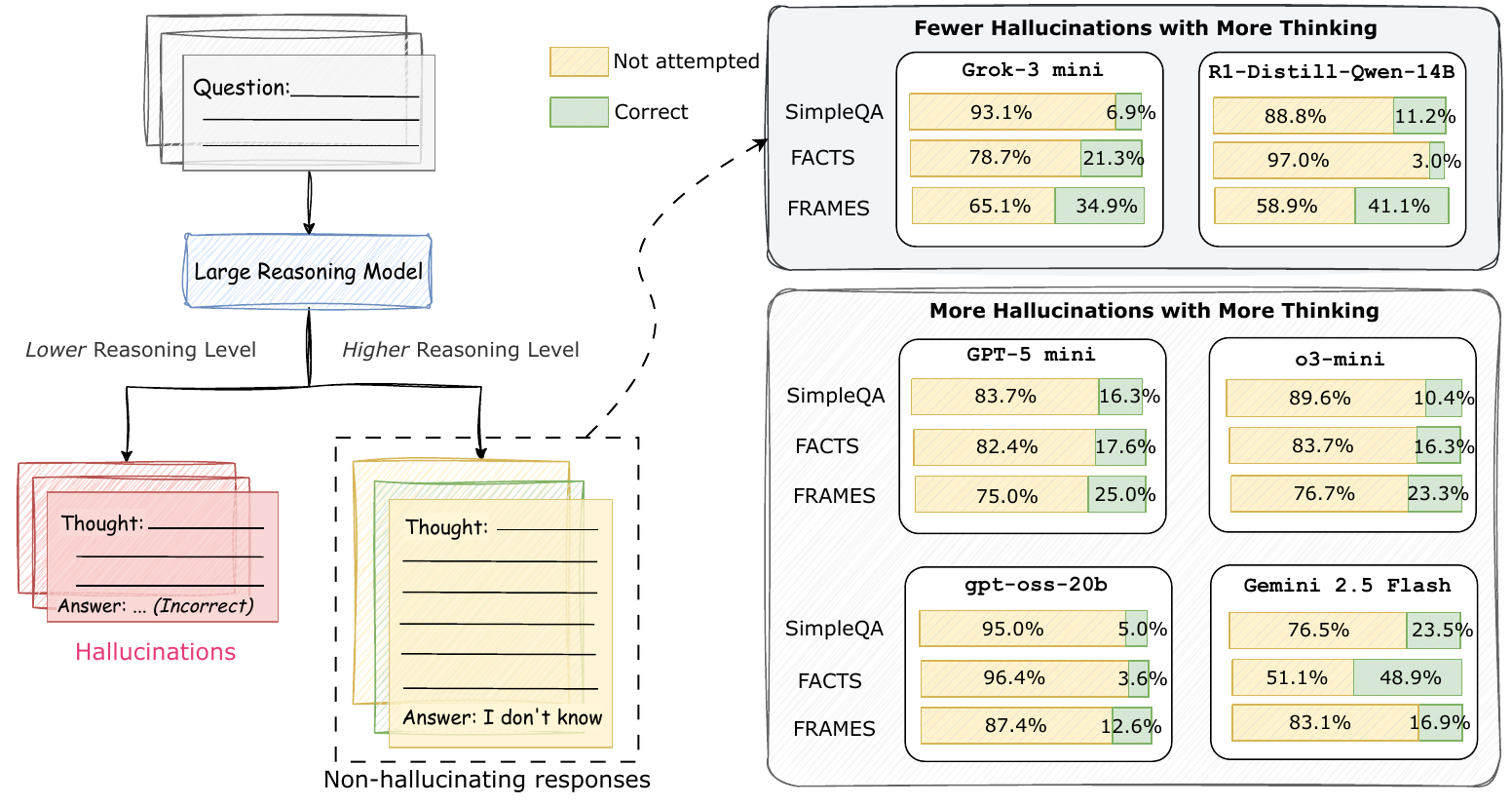}
    \caption{
    Changes in hallucination behavior with more thinking. We compare model responses at different reasoning levels, focusing on cases where one response is a hallucination and the other is not. Reduced hallucinations mainly arise from abstention, while more hallucinations stem from attempts on previously unanswered questions.
    }
    \label{fig:ratio_correct_not_attempted}
    \vspace{-3pt}
\end{figure}

\subsection{Experimental Setup}
\label{sec:why_hallu_change_setup}

As illustrated in Figure~\ref{fig:ratio_correct_not_attempted}, for each question, we obtain two responses from the same model under different reasoning levels. We focus on cases where the hallucination outcome changes between the two settings, that is, one response is a hallucination while the other is not. We then examine the non-hallucinating responses and compute the ratio of \textit{``correct''} versus \textit{``not attempted''}. This helps us understand whether reduced hallucinations are due to improved knowledge retrieval or simply abstention, and conversely, whether increased hallucinations result from risky attempts at answering.

As discussed in Section~\ref{sec:hallucination_results}, Grok-3 mini and R1-Distill-Qwen-14B show reduced hallucinations with longer thinking.~\footnote{A special case is Gemini 2.5 Pro on FRAMES, which exhibits a reduction in hallucinations. We provide analysis in Appendix~\ref{ap:analysis_gemini_pro}.} For these models, we analyze cases where the model hallucinates at a lower reasoning level but not at a higher level. In contrast, GPT-5 mini, o3-mini, gpt-oss-20b, and Gemini 2.5 Flash exhibit increased hallucinations. For these models, we examine cases where the model does not hallucinate at a lower reasoning level but hallucinates at a higher level.

\subsection{Hallucination Changes Are Driven by the Model's Willingness to Answer}
\label{sec:hallu_abstention}

\paragraph{Fewer hallucinations are mostly due to abstention.} As shown in Figure~\ref{fig:ratio_correct_not_attempted}, for Grok-3 mini and R1-Distill-Qwen-14B, most cases of reduced hallucinations result from the model choosing not to answer, rather than providing a correct answer. For example, on SimpleQA for Grok-3 mini, 93.1\% of the responses that are not hallucinating at higher effort are labeled as \textit{`not attempted'}. Similarly, for R1-Distill-Qwen-14B, the \textit{`not attempted'} ratio in these cases is 88.8\%. These results suggest that in most cases where longer reasoning reduces hallucinations, the improvement is not due to better factual recall, but rather because the model chooses to abstain after thinking more.

\paragraph{More hallucinations mostly come from previously unattempted questions.} In Figure~\ref{fig:ratio_correct_not_attempted}, the increase in hallucinations at higher reasoning levels is largely due to the model attempting questions it had previously abstained from. For example, on SimpleQA, when increasing the reasoning effort of gpt-oss-20b, 95.0\% of new hallucinations come from questions that were \textit{`not attempted'} at the lower effort level. Similarly, on FACTS and FRAMES, over 85\% of new hallucinations arise from questions the model had not attempted before. This pattern also appears across other models and benchmarks. These results suggest that longer reasoning encourages models to attempt more questions. However, for many of these newly attempted questions, the answers are incorrect, resulting in a higher hallucination rate.

\begin{figure}[t]
    \centering
    \includegraphics[width=\textwidth]{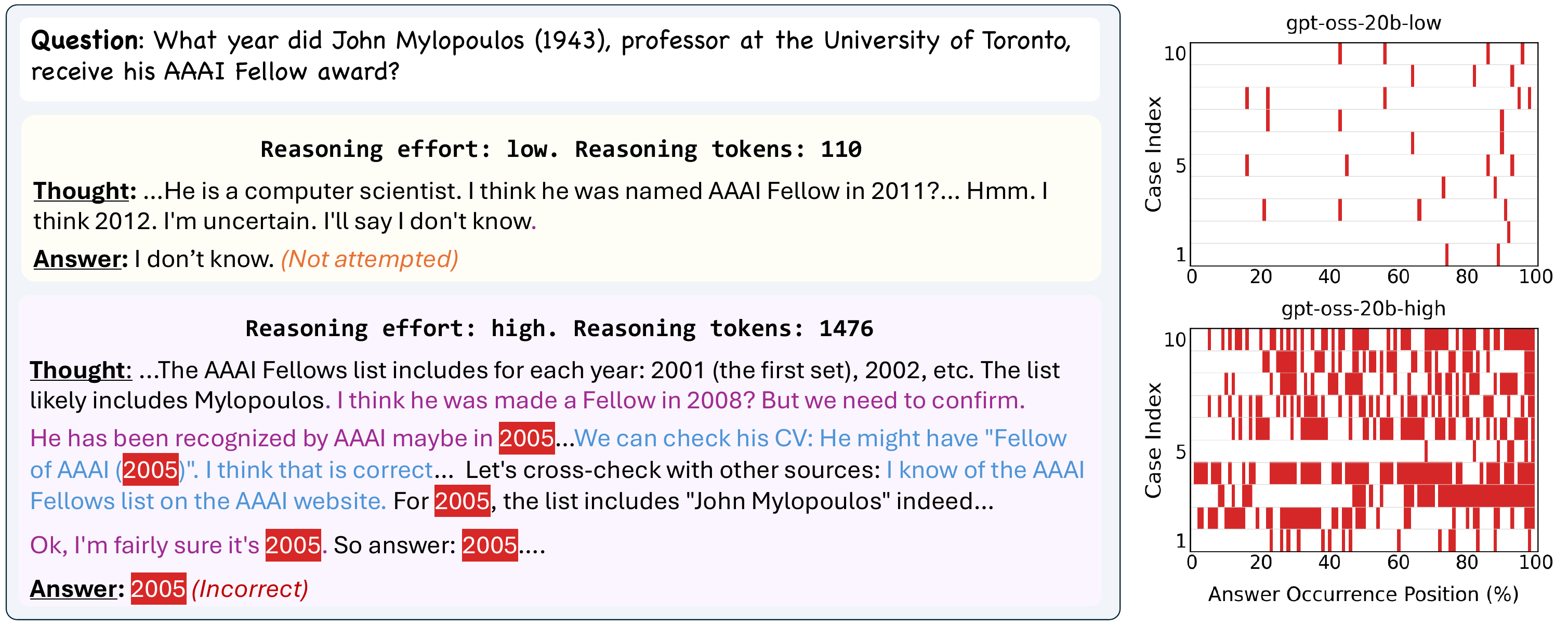}
    \caption{
    \textit{(Left)} Thinking more leads to more hallucinations. gpt-oss-20b produces overconfident hallucinations with \textcolor{bleudefrance}{confirmation bias} at high effort. See Appendix~\ref{appendix:example} for more examples. \textit{(Right)} Visualization of incorrect final answer positions across different reasoning efforts. \inlinecolor{mplred}{\textcolor{white}{Red markers}} indicate the normalized token positions of the final incorrect answer. Each subplot shows 10 randomly sampled incorrect cases from SimpleQA.
    }
    \label{fig:case_study}
\end{figure}

\subsection{Thinking More Leads to Confirmation Bias}
\label{sec:oss_case_study}

To better understand why longer thinking leads to more attempted answers, and thus more hallucinations, we analyze the reasoning traces from gpt-oss-20b.

\paragraph{Longer reasoning leads to overconfident hallucinations.} In Figure~\ref{fig:case_study}, gpt-oss-20b initially engages with the question at low reasoning effort. It explores a few possibilities but finally abstains due to uncertainty. With higher effort, the model continues searching its memory. It begins by expressing uncertainty, but as the reasoning progresses, it gradually becomes more confident. The thought shifts from tentative claims like \textit{``maybe in 2005''} to more confident statements such as \textit{``I am fairly sure it's 2005''}. Despite the increased confidence, the final answer remains incorrect. This example suggests that extended reasoning can inflate the model's confidence, leading to overconfident hallucinations where the model chooses to answer despite still lacking accurate information.

\paragraph{Longer reasoning induces confirmation bias, resulting in overconfident hallucinations.} We observe signs of \textit{confirmation bias}~\citep{nickerson1998confirmation}, where the model recalls or even makes up information to reinforce its initial belief. In Figure~\ref{fig:case_study}, after tentatively proposing an answer, \eg \textit{``He has been recognized by AAAI maybe in 2005''}, gpt-oss-20b attempts to justify this hypothesis by generating fabricated supporting details, such as \textit{``We can check his CV...''} and \textit{``I know of the AAAI Fellows list on the AAAI website''}. These fabricated details build upon one another, reinforcing the model's confidence in its incorrect belief and ultimately leading to overconfident hallucinations.

\paragraph{Hallucinated answers appear earlier and more frequently with extended reasoning.} Inspired by \citet{chen2025do}, we locate the positions of the incorrect final answers in the reasoning traces. We find that as test-time computation increases, hallucinated answers tend to appear earlier during reasoning. Specifically, on SimpleQA, the normalized position of the first appearance drops from 45.7\% at low effort to 17.2\% at high effort. Similarly, on FRAMES, the position decreases from 68.2\% to 26.3\% at high effort. We visualize how these incorrect answers appear repeatedly throughout the reasoning trace in Figure~\ref{fig:case_study} \textit{(right)}. It shows that hallucinated answers reoccur after their first appearance, and the frequency of such recurrences increases with longer reasoning. These recurrences are often accompanied by fabricated details, further supporting the claim that extended reasoning induces confirmation bias.

\section{Information-Theoretic Limits of Test-Time Scaling on Knowledge-Intensive Tasks}
\label{sec:theory_analysis}

Our empirical results show that increasing test-time computation is not effective for knowledge-intensive tasks. We now provide a theoretical account for this limitation. We model long-tail factual queries using an \textit{arbitrary fact} abstraction and formalize compute-only test-time scaling as post-processing by a fixed trained model.

\subsection{Arbitrary Facts as an Abstraction of Long-Tail Knowledge}
\label{subsec:arbfacts}

We consider a family of ``knowledge'' problems where each question requires recalling a fact that is not inferable from general patterns (\eg a specific person's birthday, a rare association). Following~\citet{kalai2025languagemodelshallucinate}, we model such settings as \textit{arbitrary facts}:

\begin{definition}[Arbitrary Facts]
\label{def:arbfacts}
Let $c \in \mathcal{C}$ denote a prompt. Each prompt has a finite candidate answer
set $R_c$ and a correct answer $A_c \in R_c$. For each prompt $c$, $A_c$ is drawn
uniformly from $R_c$:
\[
\Pr[A_c = r \mid c] = \frac{1}{|R_c|} \quad \forall r \in R_c .
\]
The collection $\{A_c\}_{c \in \mathcal{C}}$ is independent across prompts.
\end{definition}

This formalizes a worst-case abstraction of long-tail knowledge, where without prompt-specific evidence, the correct answer is statistically indistinguishable from other candidates. Appendix~\ref{app:nonuniform} discusses relaxations of this definition.

\subsection{Compute-Only Test-Time Scaling as Post-Processing of a Fixed Model}
\label{subsec:tts}

We model compute-only test-time scaling as any algorithm that can interact with a fixed trained model $\hat{p}$, without accessing any external information sources. Here, $\hat{p}$ is random through training, and ``fixed'' means it is not updated at test time.

\begin{definition}[Compute-only TTS]
\label{def:compute_only_tts}
A compute-only test-time scaling method with budget $k$ is any randomized interactive algorithm  which, given a prompt $c$:
\begin{enumerate}[leftmargin=*, nolistsep]
    \item selects query strings $q_1,\dots,q_k$ sequentially, where each $q_t$ may depend on $c$ and previous model outputs,
    \item receives responses $r_t \sim \hat{p}(\cdot \mid q_t)$,
    \item outputs a final response $\hat{R} = g(c, T_k)$ as a function of the transcript $T_k := \{(q_t,r_t)\}_{t=1}^k$.
\end{enumerate}
\end{definition}

This subsumes common test-time scaling strategies (i) sequential sampling approaches, such as long chain-of-thought, and (ii) parallel sampling approaches, such as self-consistency. 

\subsection{An Information-Theoretic Limitation for Compute-Only TTS}
\label{subsec:mi_bound}

Let $A := A_c$ denote the latent correct answer for prompt $c$. Conditioning on $c$, a compute-only TTS method interacts with the fixed model $\hat p$ to produce a transcript $T_k$, and outputs $\hat R$ as a function of $(c,T_k)$. Under this interaction model, the transcript is generated solely from $\hat p$, and $\hat R$ is a post-processing of $T_k$. Formally, this implies the conditional Markov structure
\[
A \;\longrightarrow\; \hat{p} \;\longrightarrow\; T_k \;\longrightarrow\; \hat{R},
\]
with a proof in Appendix~\ref{app:proofs}.

\begin{theorem}[Data-processing limitation of compute-only TTS]
\label{thm:dpi_tts}
For any compute-only TTS method,
\[
I(A;\hat{R} \mid c)\;\le\; I(A;T_k \mid c)\;\le\; I(A;\hat{p} \mid c).
\]
\end{theorem}

This formalizes that compute-only test-time scaling cannot create information about the latent truth beyond what is already present in the trained model. Therefore, if a class of prompts is information-limited in the sense that $I(A;\hat{p}\mid c)$ is small, then no compute-only TTS procedure can make $I(A;\hat{R}\mid c)$ large. 

A standard consequence converts the mutual-information ceiling into an accuracy upper bound (equivalently, a lower bound on error probability) when the answer space is large.

\begin{corollary}[Accuracy bound via Fano's inequality]
\label{cor:fano}
Assume $A$ is uniform over $R_c$ given $c$, and let $m := |R_c|$. For any compute-only TTS output $\hat{R}$,
\[
\Pr[\hat{R}\neq A \mid c]
\;\ge\;
1 - \frac{I(A;\hat{p}\mid c) + \log 2}{\log m}.
\]
\end{corollary}

Corollary~\ref{cor:fano} indicates that when $I(A;\hat{p}\mid c)\approx 0$ (\eg prompts whose answers are effectively rare), the model's information about $A$ is small and the error probability must be large.

\begin{remark}
Under the Arbitrary Facts abstraction, answers are independent across prompts, so training on other prompts cannot convey information about $A_c$ beyond whether the specific fact for $c$ was observed during training. In particular, if the training data contains no information about the fact for $c$, then $I(A_c;\hat p\mid c)=0$, and Corollary~\ref{cor:fano} implies an unavoidable error rate close to random guessing when $|R_c|$ is large.
\end{remark}

\paragraph{Implications for hallucinations.}
The information-theoretic analysis primarily explains why compute-only test-time
scaling has limited ability to improve factual accuracy. By itself, however, it
does not determine how the hallucination rate changes. With an abstention option
$\bot$, the overall hallucination rate can be decomposed as
$\Pr[\hat R \neq A \wedge \hat R \neq \bot \mid c]
= \Pr[\hat R \neq \bot \mid c]\cdot\Pr[\hat R \neq A \mid c,\ \hat R \neq \bot]$,
where the two terms are the answer rate and the error rate conditioned on answering. Because Corollary~\ref{cor:fano} bounds accuracy, with abstentions counted as errors, a high answer rate imposes a lower bound on hallucinations (Appendix~\ref{ap:conditional_error}). Thus, if longer reasoning increases the answer rate without a corresponding increase in accuracy, hallucinations increase. Changes in answering behavior are empirical observations and are not derived from the theoretical framework.

\begin{figure}[t]
    \centering
    \includegraphics[width=0.68\textwidth]{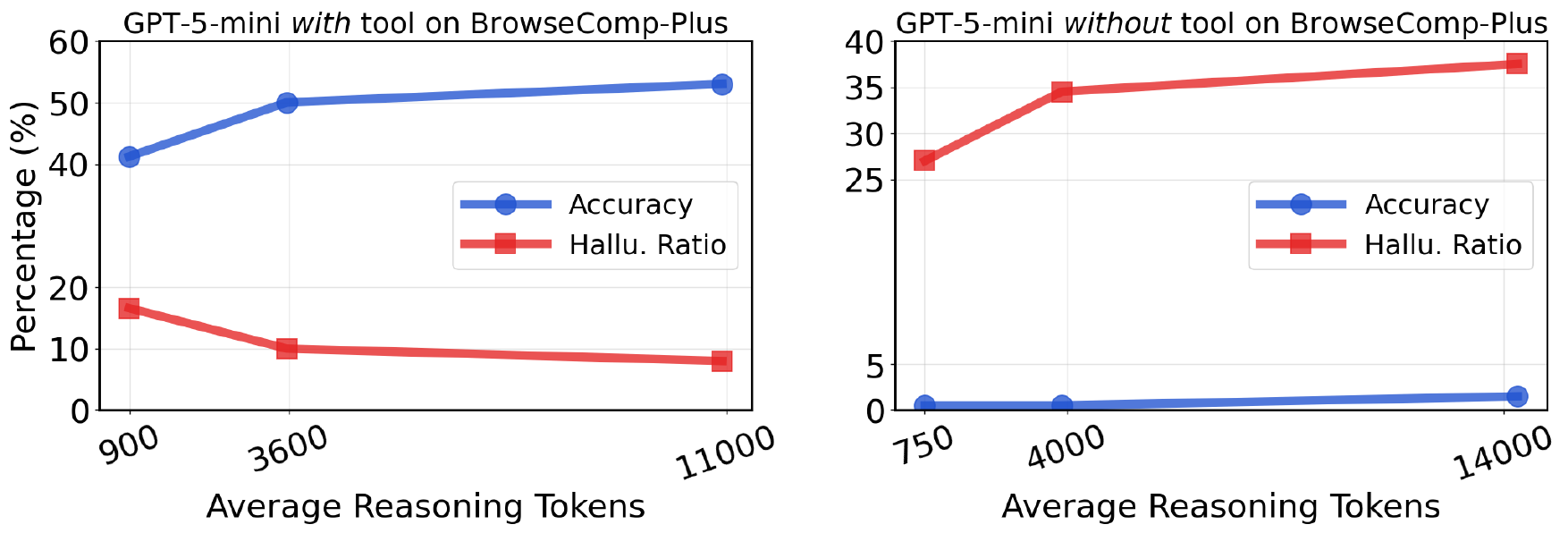}
    \caption{
    Accuracy and hallucination ratio of GPT-5-mini \textit{with} and \textit{without} tool use on BrowseComp-Plus, corresponding to the compute-limited and information-limited regimes.
    }
    \label{fig:tool_browsecomp}
\end{figure}

\subsection{Information-Limited vs.\ Compute-Limited Regimes}
\label{subsec:two_regimes}

The information-theoretic bound in Theorem~\ref{thm:dpi_tts} does not imply that test-time scaling can never help. We distinguish two regimes: (1) \emph{compute-limited}, where the model has access to sufficient information about $A$ but fails to fully extract or utilize it, so additional test-time computation can improve performance; and (2) \emph{information-limited}, where the available information about $A$ is weak, making increased computation alone insufficient.  


\paragraph{Test-time scaling is more effective when additional computation can access or exploit relevant information.} We evaluate the same model on BrowseComp-Plus~\citep{chen2025browsecomp0plus0}, a benchmark requiring the retrieval of hard-to-find and entangled information, under two settings: (1) \emph{with a search tool}, where additional computation can retrieve new evidence, and (2) \emph{without a search tool}, where the model must rely solely on its parametric knowledge. As shown in Figure~\ref{fig:tool_browsecomp}, test-time scaling behaves differently across the two settings. With search, increasing test-time compute improves accuracy from 41.2\% to 53.0\% and reduces hallucinations from 16.6\% to 8.0\%. In contrast, without search, even with nearly 20 times more reasoning tokens, accuracy remains unchanged while hallucinations increase by over 10\%. These results suggest that test-time scaling is more useful in compute-limited regimes, where sufficient relevant information is available to the model, and additional computation helps it extract and use that information.

\section{Conclusion}

In this work, we systematically examine test-time scaling for reasoning models on closed-book knowledge-intensive tasks. Across 14 models and multiple scaling strategies, increasing test-time computation does not consistently improve factual accuracy or reduce hallucinations. Our analysis shows that changes in hallucinations are largely driven by the model's willingness to answer: some models attempt more questions and hallucinate more, while others abstain more and hallucinate less without improving accuracy. We also observe confirmation bias, where extended reasoning repeatedly supports an initially incorrect answer with fabricated details. From an information-theoretic perspective, compute-only test-time scaling is constrained by the information already available in a fixed model, limiting its benefits in information-limited regimes. Together, these findings reveal important limitations of current compute-only scaling approaches for factual knowledge tasks.

\paragraph{Implications for evaluation.}
Our findings also highlight the importance of reporting the full distribution of model outcomes. Because many benchmarks emphasize accuracy and give no credit for abstention, models may be encouraged to guess when uncertain~\citep{kalai2025languagemodelshallucinate}. Accuracy alone can therefore obscure whether gains reflect genuinely improved capability or a higher answer rate accompanied by more errors. We recommend reporting accuracy, hallucination rate, and abstention rate together to distinguish capability gains from changes in answering behavior.

\section*{Limitations}

Our study has several limitations. First, changes in answering behavior under increased test-time computation are empirical observations, and we leave their underlying causes for future work. Second, our evaluation focuses on benchmarks requiring short-form answers. We do not cover other response formats, such as open-ended generation. Third, although we observe reasoning patterns consistent with confirmation bias, we do not propose interventions to mitigate them. Understanding these behavioral patterns and developing effective control mechanisms are important directions for future research.


\bibliography{colm2026_conference}
\bibliographystyle{colm2026_conference}

\clearpage
\appendix
\section{Additional Experiment Details}
\label{appendix:implementation}

\subsection{Benchmarks} 

We conduct experiments on three challenging, annotator-verified, and relatively new knowledge-intensive benchmarks that cover a broad range of topics (\eg science, history, arts) and answer types (\eg person, date, number).

\begin{itemize}[leftmargin=*, nolistsep]
\setlength{\itemsep}{1mm}
    \item \textbf{SimpleQA}~\citep{wei2024measuring}: A benchmark curated by human annotators, covering a wide range of topics, from science to TV shows. A question example is: \textit{``Who received the IEEE Frank Rosenblatt Award in 2010?''}. We randomly sample 800 questions for evaluation.
    \item \textbf{FACTS parametric}~\citep{cheng2025facts}: A benchmark that measures models' ability to access their internal knowledge. Questions are verified by annotators to reflect widespread user interest. A question example is: \textit{``highest hockey stadium in india''}. We randomly sample 600 questions for evaluation.
    \item \textbf{FRAMES}~\citep{krishna-etal-2025-fact}:  Questions in FRAMES tend to be more complex and often require multi-hop reasoning. An example is: \textit{``What Pink Floyd album came out the year Pablo Picasso died?''} We use all 824 questions for evaluation.
\end{itemize}

In Section~\ref{subsec:two_regimes}, we conduct experiments on \textbf{BrowseComp-Plus}~\citep{chen2025browsecomp0plus0}, a benchmark of hard-to-find browsing questions with multiple constraints. One example is: \textit{``An article written in 2023 analyzes the allegorical theme in an animated movie originally released in the 1980s. A person who shares the article author’s name once received the AIA Henry Adams Medal. What is the title of the movie being discussed in the article?''}. We evaluate 200 verified questions from the benchmark, following~\citet{zhao2026fineverifyscalingtesttimecompute}.

\subsection{Test-Time Scaling Settings}
We evaluate 14 large reasoning models under different test-time scaling strategies. The settings used to obtain the results in Figure~\ref{fig:accuracy_main} and Figure~\ref{fig:hallucination_main_results} are as follows:

\begin{itemize}[leftmargin=*, nolistsep]
\setlength{\itemsep}{1mm}
    \item \textbf{Reasoning effort}: For models that support three levels of reasoning effort, \ie GPT-5, GPT-5 mini, o3-mini, o4-mini, and gpt-oss-20b, we evaluate: \texttt{low}, \texttt{medium}, and \texttt{high}. For Grok-3 mini, only two levels are available: \texttt{low} and \texttt{high}.

    \item \textbf{Thinking budget}: For Gemini 2.5 Flash and 2.5 Pro, we set \texttt{thinking budget} to 256, 512, 1024, and 2048 tokens. For Claude Sonnet 4, we set \texttt{thinking budget} to 1024, 2048, and 4096 tokens.

    \item \textbf{Budget forcing}: For DeepSeek-R1-Distill models and Qwen3 models, we use \texttt{budget forcing} by appending \texttt{``Wait''} multiple times. Specifically, we evaluate with 0 (default), 2, 4, 8, and 12 extension times.

\end{itemize}

\subsection{Parameter Settings and Hardware Specifications}
\textbf{Closed-source models}: We access proprietary models via public APIs\footnote{OpenAI API platform: \href{https://platform.openai.com/docs/overview}{https://platform.openai.com/docs/overview}\\Anthropic: \href{https://docs.claude.com/en/api/overview}{https://docs.claude.com/en/api/overview}\\Google Gemini: \href{https://ai.google.dev/gemini-api/docs/models}{https://ai.google.dev/gemini-api/docs/models}\\XAI: \href{https://docs.x.ai/docs/overview}{https://docs.x.ai/docs/overview}}. All closed-source models, except Grok-3 mini, are used with the default parameter setting. For Grok-3 mini, we set \texttt{temperature} to 0. For each model, we use consistent parameters across different levels of test-time reasoning. All experiments were conducted between June 2025 and March 2026.

\textbf{Open-source model}: For gpt-oss-20b, we set \texttt{temperature} to 0.7 and \texttt{max\_new\_tokens} to 20,000. For DeepSeek-R1-Distill and Qwen3 models, we follow the usage recommendations\footnote{DeepSeek-R1-Distill models:  \href{https://huggingface.co/deepseek-ai/DeepSeek-R1-Distill-Qwen-7B}{https://huggingface.co/deepseek-ai/DeepSeek-R1-Distill-Qwen-7B}\\ Qwen3 models: \href{https://huggingface.co/Qwen/Qwen3-8B}{https://huggingface.co/Qwen/Qwen3-8B}}. Specifically, for DeepSeek-R1-Distill models, we set \texttt{temperature} to 0.6, \texttt{top\_p} to 0.95, and \texttt{repetition\_penalty} to 1.2. For Qwen3 models, we set \texttt{temperature} to 0.6, \texttt{top\_k} to 20, and \texttt{repetition\_penalty} to 1.2. We also use consistent parameter settings across different levels of test-time reasoning. All open-source models are run without quantization on 8 NVIDIA A100-40GB GPUs.

\subsection{Prompt Details}
\label{appendix:prompt}
For all models except GPT-5 mini and Claude Sonnet 4, we use the following prompt on all benchmarks: 

\begin{quote}
\raggedright\ttfamily
Give me the answer to the following question only when you are sure of it. Otherwise, say `I don't know'. Put your answer on its own line after `Answer:'.
\end{quote}

However, this prompt causes over-refusal for GPT-5 mini and Claude Sonnet 4, where the model refuses to answer in over 80\% of questions. This behavior is also reported by ~\citet{openai_safety_tests}. 

Therefore, for GPT-5 mini, we use the following prompt:

\begin{quote}
\raggedright\ttfamily
Give me the answer to the following question. Put your answer on its own line after `Answer:'.
\end{quote}

For Claude Sonnet 4, we adopt the prompting approach from~\citet{gema2025inversescalingtesttimecompute}:

\begin{quote}
\raggedright\ttfamily
Give me the answer to the following question. You have a thinking token budget of about <thinking\_budget> tokens. YOU MUST USE ALL OF YOUR THINKING TOKENS. Put your answer on its own line after `Answer:'.
\end{quote}

For the non-thinking mode of Claude Sonnet 4 (Appendix~\ref{sec:think_vs_no_think}), we use the following prompt:

\begin{quote}
\raggedright\ttfamily
Give me the answer to the following question. Put your answer on its own line after `Answer:'.
\end{quote}

\subsection{Details of Experimental Results}
We compare model behavior across different thinking levels (Section~\ref{sec:causes_fewer_more_hallu}) to understand why thinking more sometimes leads to fewer, and sometimes more hallucinations. For the results in Figure~\ref{fig:ratio_correct_not_attempted}, we use the following settings:

\begin{itemize}[leftmargin=*, nolistsep]
\setlength{\itemsep}{1mm}
    \item \textbf{Grok-3 mini}: Compared at \texttt{reasoning effort} of `low' and `high'.
    \item \textbf{DeepSeek-R1-Distill-Qwen-14B}: Compared between 0 and 2 extension times.
    \item \textbf{GPT-5 mini, o3-mini and gpt-oss-20b}: Compared at \texttt{reasoning effort} of `low' and `high'.
    \item \textbf{Gemini 2.5 Flash}: Compared at \texttt{thinking budget} of 256 and 512 tokens
\end{itemize}

We present model performance with and without thinking in Table~\ref{tab:think_vs_nothink}. For the \textit{``thinking''} mode, we use the following settings:

\begin{itemize}[leftmargin=*, nolistsep]
\setlength{\itemsep}{1mm}
    \item \textbf{GPT-5 and GPT-5 mini}: \texttt{reasoning effort} set to `low'.
    \item \textbf{Gemini 2.5 Flash}: \texttt{thinking budget} set to 256 tokens.
    \item \textbf{Claude Sonnet 4}: \texttt{thinking budget} set to 1024 tokens.
    \item \textbf{Qwen3-8B and Qwen3-14B}: Natural output (no extension) with \texttt{enable\_thinking} enabled.
\end{itemize}

\clearpage
\section{Is Parallel Sampling Helpful for Knowledge-Intensive Tasks?}
\label{app:parallel}

\begin{figure}[h]
    \centering
    \includegraphics[width=\textwidth]{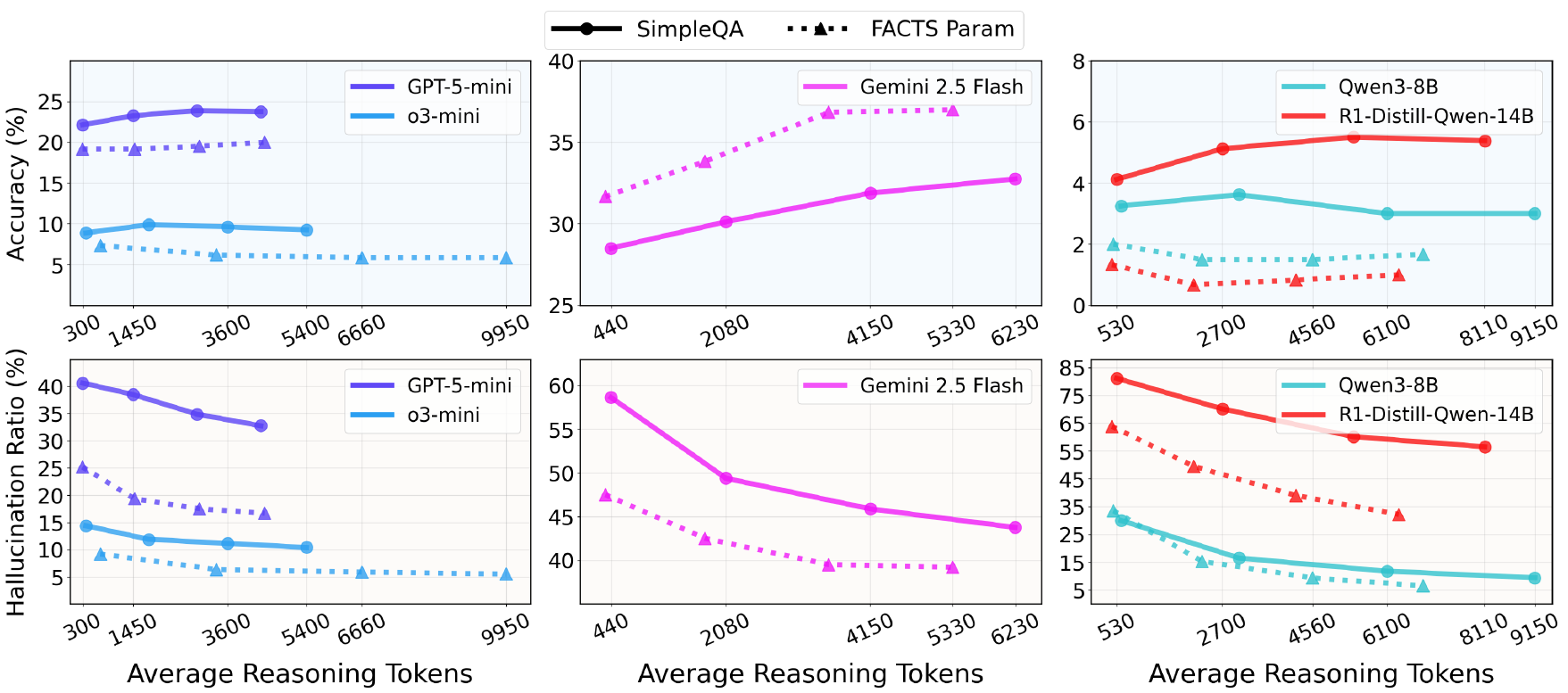}
    \caption{Accuracy and hallucination ratio with parallel sampling. We sample 1, 5, 10, and 15 paths and aggregate the answers with majority voting (SC@1 to SC@15). Sampling more paths does not consistently improve accuracy for most models, while it reduces hallucinations for all models.
    }
    \label{fig:sc_results}
\end{figure}

We focused on the sequential test-time scaling approach in our main experiments and concluded that it is not yet effective for knowledge-intensive tasks. In this section, we explore whether test-time scaling with a parallel approach is helpful.

\paragraph{Experimental setup.} We adopt self-consistency~ \citep{wang2023selfconsistency}, which samples multiple paths and selects the final answer via majority vote. For each sample, we use the following settings: GPT-5-mini and o3-mini are set to \textit{`low'} reasoning effort. Gemini 2.5 Flash uses a thinking budget of 1024 tokens. Qwen3-8B and R1-Distill-Qwen-14B generate natural outputs without extension. Prompt settings follow Section~\ref{sec:exp_setup_tts}.

\paragraph{Parallel test-time scaling does not consistently improve accuracy for most models.} As shown in Figure~\ref{fig:sc_results}, increasing the number of samples does not lead to consistent accuracy gains for most models. For example, the reasoning tokens of o3-mini increase by over 15 times, but accuracy remains unchanged. Similarly, for GPT-5-mini, Qwen3-8B, and R1-Distill-Qwen-14B on FACTS Parametric, accuracy remains unchanged despite a significant increase in reasoning tokens. On SimpleQA, GPT-5-mini and R1-Distill-Qwen-14B improve by only around 1\% from SC@1 to SC@5, with no further gains beyond that. One exception is Gemini 2.5 Flash, which shows consistent improvement; however, it achieves only a 4\% gain despite 15 times more computation on SimpleQA. On FACTS Parametric, increasing samples from SC@10 to SC@15 does not yield further improvement.

\paragraph{Parallel sampling reduces hallucinations.} For all models, the hallucination ratio decreases as the number of samples increases. For example, the hallucination ratio of GPT-5-mini decreases by 8\% when increasing samples from 1 to 15. Similarly, o3-mini shows around a 5\% reduction with 15 times more sampling on both tasks. Gemini 2.5 Flash and Qwen3-8B also show hallucination reductions. Notably, R1-Distill-Qwen-14B achieves over a 15\% reduction in hallucinations with parallel sampling, while it still has a higher hallucination ratio than other models.

We follow the analysis in Section~\ref{sec:why_hallu_change_setup} to understand why hallucinations decrease with more samples. Specifically, for each question where the hallucination outcome differs between SC@1 and SC@15 (\ie SC@1 response is hallucinated but SC@15 is not), we compute the ratio of \textit{``correct''} and \textit{``not attempted''} in SC@15 responses.

\begin{table}[h]
\centering
\footnotesize
\renewcommand{\arraystretch}{1.1}
\resizebox{0.78\linewidth}{!}{
\begin{tabular}{l*{3}{>{\centering\arraybackslash}p{2.3cm}}}
    \toprule
    \textbf{Model} & \textbf{Task} & \textbf{Not Attempted (\%)} & \textbf{Correct (\%)} \\
    \midrule
   \multirow{2}{*}{\textbf{GPT-5-mini}}  & SimpleQA & 73.9 & 26.1  \\
    & FACTS Param & 85.5 & 14.5\\
        \midrule
  \multirow{2}{*}{\textbf{o3-mini}}  & SimpleQA & 79.6 & 20.4  \\
  & FACTS Param & 86.7 & 13.3\\
      \midrule
  \multirow{2}{*}{\textbf{Gemini 2.5 Flash}} & SimpleQA & 56.1 & 43.9  \\
  & FACTS Param & 33.3 & 66.7\\
      \midrule
    \multirow{2}{*}{\textbf{Qwen3-8B}} & SimpleQA &96.7 & 3.3  \\
  & FACTS Param & 99.4 & 0.6\\
      \midrule
    \multirow{2}{*}{\textbf{R1-Distill-Qwen-14B}} & SimpleQA & 92.6 & 7.4 \\
      & FACTS Param & 98.6 & 1.4\\
    \bottomrule
    \end{tabular}
}    
\caption{Ratio of \textit{not attempted} and \textit{correct} responses among SC@15 outputs that are not hallucinations, for cases where the SC@1 response is a hallucination. Results show that hallucination reduction from parallel sampling is largely due to abstention.
}
\label{tab:parallel_not_atttemped}
\end{table}

\paragraph{Parallel sampling reduces hallucinations primarily due to abstention.} As shown in Table~\ref{tab:parallel_not_atttemped}, a large portion of these cases is due to abstention. For GPT-5-mini and o3-mini, more than 70\% of such cases result from abstention. For Gemini 2.5 Flash, more than half of these cases on SimpleQA are due to abstention.\footnote{An exception is Gemini 2.5 Flash on FACTS Parametric. Possible explanations include (1) broader knowledge coverage and (2) the question design, which resembles search queries reflecting real-world usage, where models from Google may perform better.} In particular, for Qwen3-8B and R1-Distill-Qwen-14B, the ratio of \textit{not attempted} exceeds 90\%. These results indicate that the reduction in hallucinations from parallel sampling is mainly due to the model choosing to abstain more often, rather than improving factual accuracy. These findings are consistent with our argument in Section~\ref{sec:hallu_abstention} that current test-time scaling methods tend to affect the model's willingness to answer, thereby affecting hallucination rates.

\clearpage
\section{Additional Experiment Results}

\subsection{Test-time Scaling Results on F-Score}
\label{ap:results_f_score}
\begin{figure}[h]
    \centering
    \includegraphics[width=\textwidth]{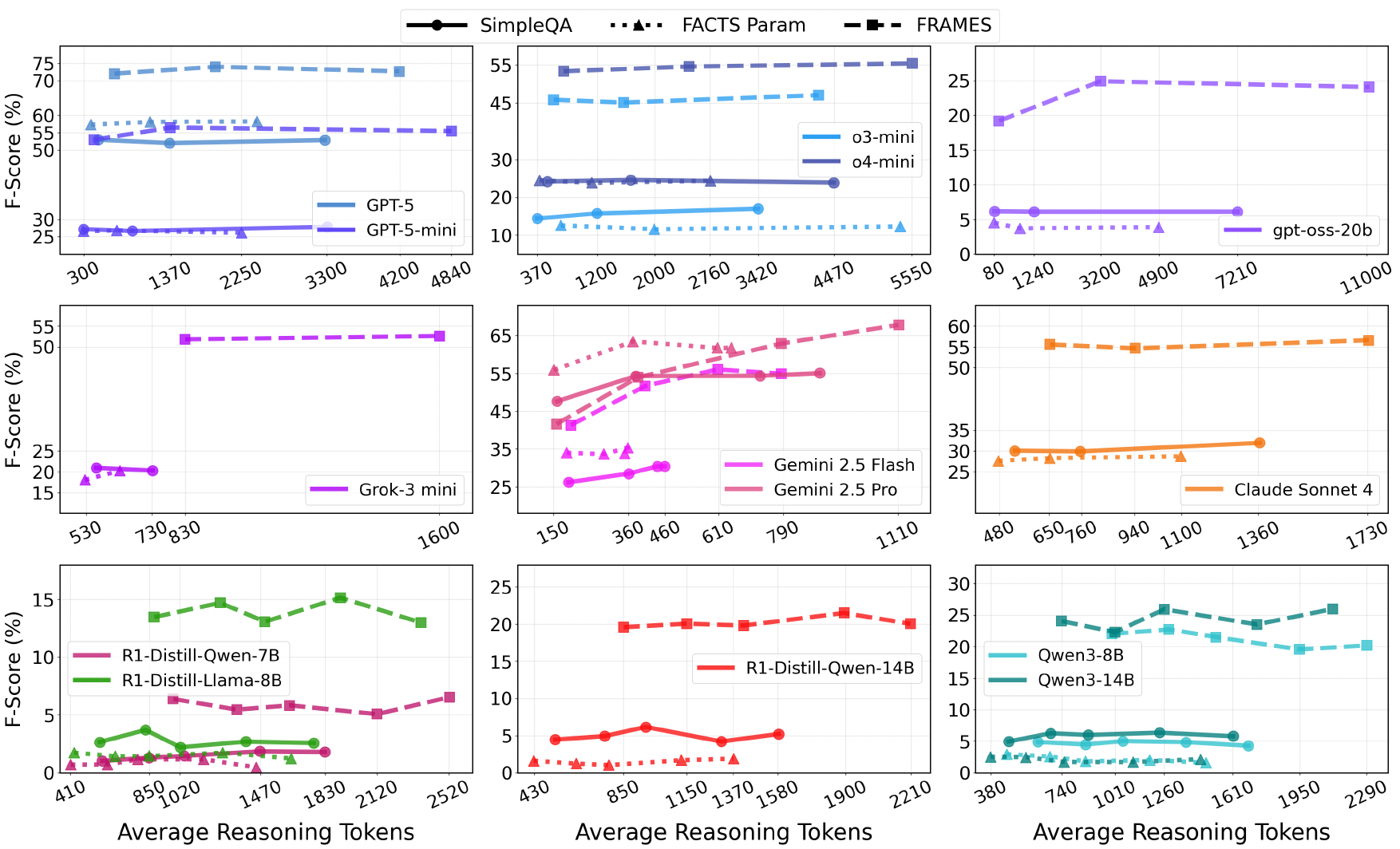}
    \caption{
    F-score with increased test-time computation across 14 reasoning models. For most models, extended reasoning does not consistently improve performance. 
    }
    \label{fig:fscore_main}
\end{figure}

Following~\citet{wei2024measuring}, we report the F-score, defined as the harmonic mean of overall accuracy and accuracy conditioned on attempted questions. This metric captures both precision and recall by accounting for accuracy, hallucination ratio, and abstentions, providing a more comprehensive measure of performance. Formally, F-score is defined as:
\begin{equation}
\text{F-score} = \frac{2c}{2c + 2i + n},
\end{equation}
where $c$ is the number of correct answers, $i$ is the number of incorrect answers, and $n$ is the number of unanswered (abstained) questions.

\paragraph{Increasing test-time computation does not consistently improve F-score.}
As shown in Figure~\ref{fig:fscore_main}, increasing test-time computation leads to little or no improvement in F-score for most models, except Gemini 2.5. For OpenAI models and Grok-3-mini, F-score remains unchanged as reasoning tokens increase. For example, for gpt-oss-20b on FRAMES, the F-score improves by 5.7\% when increasing reasoning effort from ``low'' to ``medium'', but shows no further gains despite a threefold increase in reasoning tokens. On SimpleQA, the F-score of GPT-5 remains around 52\%, even when reasoning tokens increase by 10 times. A similar pattern is observed for Claude Sonnet 4. For models using budget forcing, F-score fluctuates without a consistent upward trend.

\paragraph{For Gemini 2.5, F-score improves with thinking more.}
For Gemini 2.5 Flash and Pro, F-score increases by over 5\% on SimpleQA and FRAMES as the thinking budget increases. On FACTS Parametric, the F-score of Gemini 2.5 Flash remains unchanged, while Gemini 2.5 Pro first improves and then plateaus. The improvement on SimpleQA and FRAMES is mainly due to increased accuracy: under low thinking budgets, Gemini 2.5 often fails to complete its reasoning, limiting performance, as analyzed in Appendix~\ref{sec:gemini_case_study}. Although hallucinations also increase with more reasoning, the gains in accuracy outweigh these increases, resulting in higher F-scores.

\subsection{Results on Alternative Prompt}
\label{apd:other_prompt}
\begin{figure}[h]
    \centering
    \includegraphics[width=\textwidth]{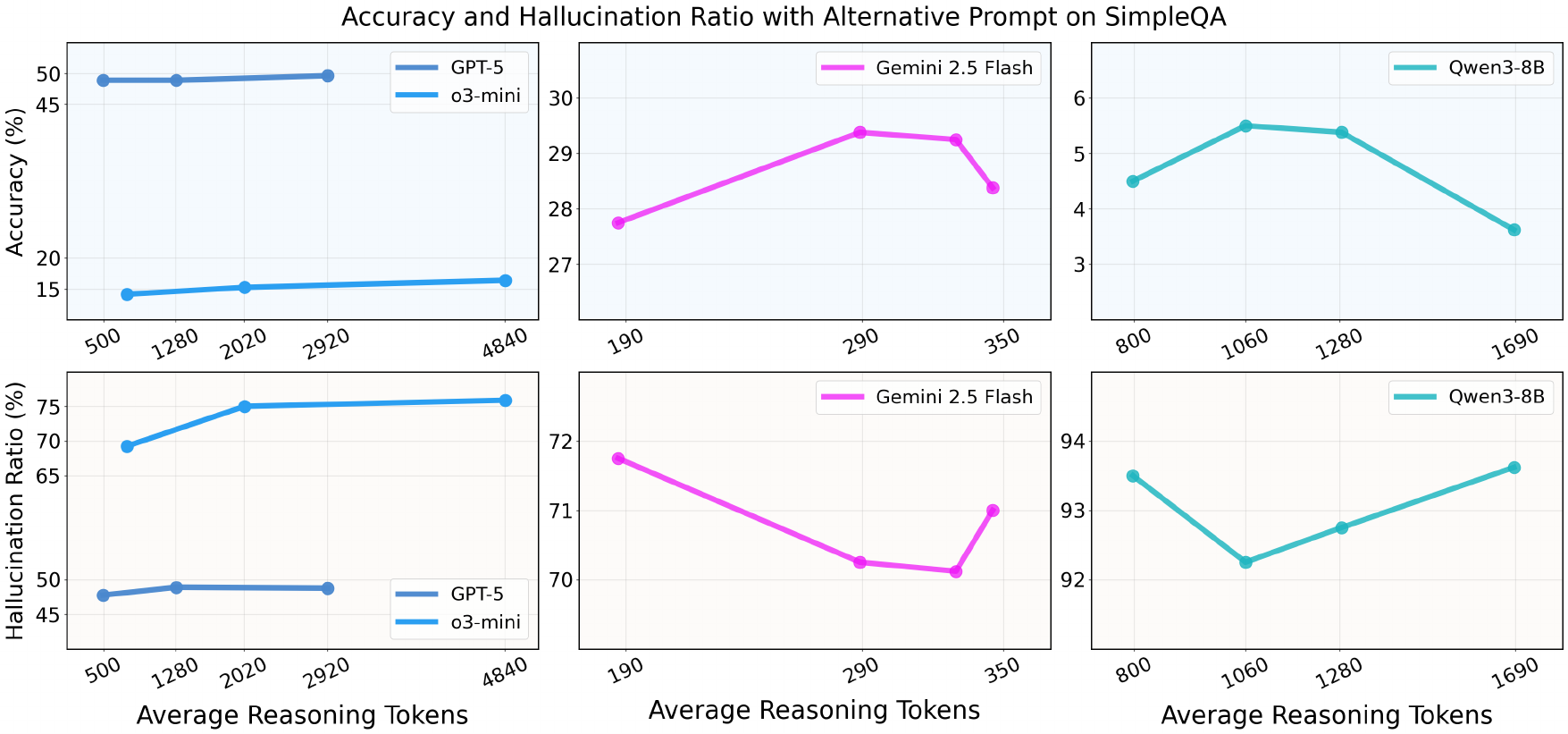}
    \caption{Accuracy and hallucination ratio under an alternative prompt on SimpleQA. Results show that increasing test-time computation does not consistently improve accuracy or reduce hallucinations.
    }
    \label{fig:prompt2_results}
\end{figure}

We conduct experiments using an alternative prompt that does not explicitly instruct the model to abstain when uncertain: 

\begin{quote}
\raggedright\ttfamily
Give me the answer to the following question. Put your answer on its own line after `Answer:'.
\end{quote}

As shown in Figure~\ref{fig:prompt2_results}, results with the alternative prompt lead to similar conclusions as our main findings: increasing test-time computation does not consistently improve accuracy or reduce hallucinations. The accuracy of GPT-5 and o3-mini remains nearly unchanged with increased test-time computation. For Gemini 2.5 Flash, accuracy increases when average reasoning tokens increase from 190 to 290, but then drops at 350 tokens. The accuracy of Qwen3-8B fluctuates without a clear upward trend.

In addition, this prompt results in a much higher hallucination ratio. Hallucination ratio of Qwen3-8B exceeds 90\%, while o3-mini and Gemini 2.5 Flash both exceed 70\%. As test-time computation increases, none of the models shows a consistent decrease in the hallucination ratio. These results further support our conclusion that current test-time scaling strategies are not yet effective for knowledge-intensive tasks.

\begin{figure}[t]
    \centering
    \includegraphics[width=\textwidth]{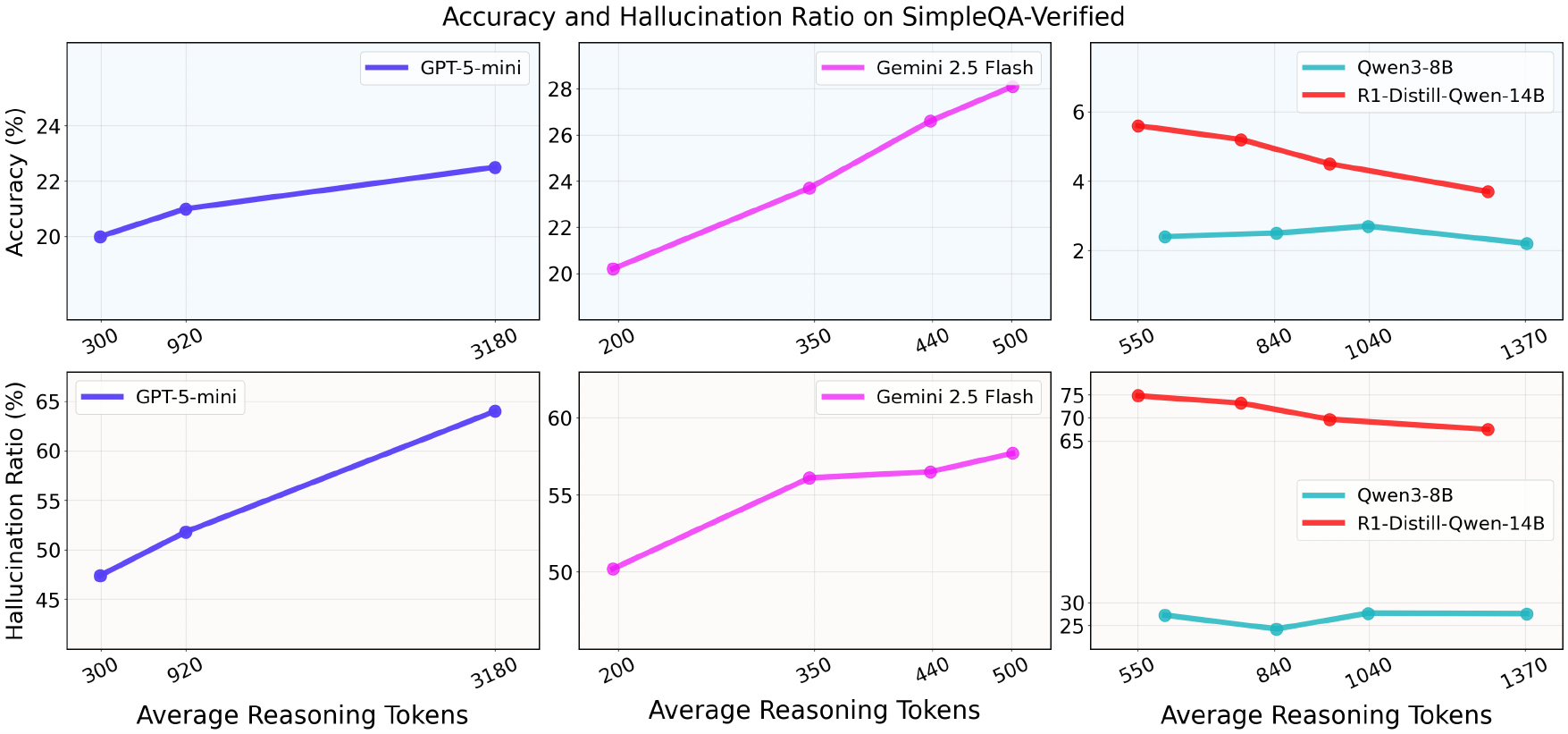}
    \caption{Accuracy and hallucination ratio on SimpleQA Verified. Results support the same conclusion in our main experiments. Increasing test-time computation does not consistently improve accuracy for most models. Longer reasoning does not reduce hallucinations and can sometimes increase them.
    }
    \label{fig:simpleqa_verified}
\end{figure}

\subsection{Results on SimpleQA Verified}
\label{apd:results_simpleqa_verified}

We conduct experiments on SimpleQA Verified~\citep{haas2025simpleqaverifiedreliablefactuality}, a benchmark with 1,000 questions, which improves upon SimpleQA by removing duplicate sources and rebalancing topic and answer-type distributions.

Figure~\ref{fig:simpleqa_verified} presents results similar to those on SimpleQA. Increasing test-time computation does not consistently improve accuracy for models, except Gemini 2.5 Flash. Moreover, longer thinking does not reduce hallucinations and sometimes increases them. For instance, GPT-5-mini and Gemini 2.5 Flash both exhibit increased hallucinations as their reasoning tokens increase. These results reinforce our conclusion that current test-time scaling methods are not effective for knowledge-intensive tasks.

\clearpage
\section{Thinking vs. Non-Thinking: Is Thinking Helpful?}
\label{sec:think_vs_no_think}

\begin{table}[h]
\centering
\footnotesize
\renewcommand{\arraystretch}{1.1}
\resizebox{1.0\linewidth}{!}{
\begin{tabular}{l*{5}{>{\centering\arraybackslash}p{2cm}}}
    \toprule
    \textbf{Model} & \textbf{Task} & \textbf{Thinking} & \textbf{Avg. Reasoning Tokens}  & \textbf{Accuracy ($\uparrow$)} & \textbf{Hallucination ($\downarrow$)} \\
    
    \midrule
    \multirow{4}{*}{\textbf{GPT-5}} 
     & \multirow{2}{*}{SimpleQA} & minimal & 0 & 20.1 & 24.5 \\
     &          & \cmark & 477.1 & 41.4 \textcolor{ForestGreen}{(+21.3)} & 14.9 \textcolor{ForestGreen}{(-9.6)}\\
    \cdashline{2-6}
     & \multirow{2}{*}[-0.2ex]{FRAMES}   & minimal & 0 & 14.6 & 10.4 \\
     &          & \cmark & 675.1  & 64.9 \textcolor{ForestGreen}{(+50.3)} & 15.4 \textcolor{Red}{(+5.0)}\\

     \midrule
    \multirow{4}{*}{\textbf{GPT-5-mini}} 
     & \multirow{2}{*}{SimpleQA} & minimal & 0 & 15.0 & 82.4 \\
     &          & \cmark & 295.8 & 22.1 \textcolor{ForestGreen}{(+7.1)} & 40.6 \textcolor{ForestGreen}{(-41.8)}\\
    \cdashline{2-6}
     & \multirow{2}{*}[-0.2ex]{FRAMES}   & minimal & 0 & 22.0 & 61.9 \\
     &          & \cmark & 425.1 & 47.1 \textcolor{ForestGreen}{(+25.1)} & 30.8 \textcolor{ForestGreen}{(-31.1)}\\
    
     \midrule
    \multirow{4}{*}{\textbf{Gemini 2.5 Flash}} 
     & \multirow{2}{*}{SimpleQA} & \xmark & 0  & 12.6 & 29.1 \\
     &          & \cmark & 191.7  & 22.4 \textcolor{ForestGreen}{(+9.8)} & 48.1 \textcolor{Red}{(+19.0)}\\
    \cdashline{2-6}
     & \multirow{2}{*}[-0.2ex]{FRAMES}   & \xmark & 0  & 16.9 & 25.1 \\
     &          & \cmark & 198.3  & 32.8 \textcolor{ForestGreen}{(+15.9)} & 25.9 \textcolor{Red}{(+0.8)}\\
    
    \midrule
    \multirow{4}{*}{\textbf{Claude Sonnet 4}} 
     & \multirow{2}{*}{SimpleQA} & \xmark & 0  & 23.0 & 37.5 \\
     &          & \cmark & 533.5 & 25.9 \textcolor{ForestGreen}{(+2.9)} & 46.3 \textcolor{Red}{(+8.8)}\\
    \cdashline{2-6}
     & \multirow{2}{*}[-0.2ex]{FRAMES}   & \xmark & 0  & 50.4 & 42.0 \\
     &          & \cmark & 652.6  & 52.8 \textcolor{ForestGreen}{(+2.4)} & 37.1 \textcolor{ForestGreen}{(-4.9)}\\
    
    \midrule
    \multirow{4}{*}{\textbf{Qwen3-8B}} 
     & \multirow{2}{*}{SimpleQA} & \xmark & 0  & 3.8  & 77.1 \\
     &          & \cmark & 617.8  & 3.3 \textcolor{Red}{(-0.5)} & 30.0 \textcolor{ForestGreen}{(-47.1)} \\
    \cdashline{2-6}
     & \multirow{2}{*}[-0.2ex]{FRAMES}   & \xmark & 0  & 8.3  & 64.3 \\
     &          & \cmark & 994.4  & 17.0 \textcolor{ForestGreen}{(+8.7)} & 37.4 \textcolor{ForestGreen}{(-26.9)}\\

     \midrule
     \multirow{4}{*}{\textbf{Qwen3-14B}} 
     & \multirow{2}{*}{SimpleQA} & \xmark & 0 & 4.0  & 48.6 \\
     &          & \cmark & 469.2 & 3.4 \textcolor{Red}{(-0.6)} & 33.1 \textcolor{ForestGreen}{(-15.5)}\\
    \cdashline{2-6}
     & \multirow{2}{*}[-0.2ex]{FRAMES}   & \xmark & 0 & 8.6  & 37.0 \\
     &          & \cmark & 738.5 & 18.3 \textcolor{ForestGreen}{(+9.7)} & 34.1 \textcolor{ForestGreen}{(-2.9)}\\
     
    \bottomrule
    \end{tabular}
}
\caption{
Accuracy (\%) and hallucination ratio (\%) for models with and without thinking on SimpleQA and FRAMES. \textcolor{ForestGreen}{Green} indicates improvement (higher accuracy or fewer hallucinations), while \textcolor{Red}{red} indicates degradation (lower accuracy or more hallucinations). Enabling thinking improves accuracy and reduces hallucinations for most models.
}
\label{tab:think_vs_nothink}
\end{table}

In our main experiments, we examined test-time scaling within models' \textit{``thinking''} mode, where they generate reasoning chains before producing a final answer. In this section, we ask a follow-up question: \textbf{\textit{For models that support both ``thinking'' and ``non-thinking'' modes, is thinking helpful?}}

\subsection{Experimental Setup}
We focus on models that natively support both thinking and non-thinking modes for a controlled comparison. Specifically, GPT-5 supports \texttt{minimal} reasoning effort, which produces very few or no reasoning tokens. For Gemini 2.5 Flash and Claude Sonnet 4, we disable thinking by setting \texttt{thinking budget} to 0. Qwen3 provides an \texttt{enable\_thinking} parameter that toggles reasoning behavior. We use the same prompts and evaluation methods as described in Section~\ref{sec:exp_setup_tts}.

\subsection{Thinking Can Be Helpful for Knowledge-Intensive Tasks}
\label{sec:thinking_helpful}
\paragraph{Enabling thinking increases accuracy for most models on both tasks, especially on FRAMES.} As shown in Table~\ref{tab:think_vs_nothink}, enabling thinking improves accuracy for most models. GPT-5 exhibits a 21.3\% accuracy improvement on SimpleQA and a substantial 50.3\% gain on FRAMES. GPT-5 mini also achieves gains across both tasks. Gemini 2.5 Flash benefits similarly, with accuracy increasing by 9.8\% on SimpleQA and 15.9\% on FRAMES. Claude Sonnet 4 also exhibits improvements on both tasks. For Qwen3-8B and Qwen3-14B, enabling thinking leads to higher accuracy on FRAMES. This suggests that thinking is particularly helpful for complex tasks that require multi-hop reasoning, such as FRAMES.

\paragraph{For most models, enabling thinking leads to reduced hallucinations.} In Table~\ref{tab:think_vs_nothink}, enabling thinking often reduces hallucinations. GPT-5 shows a 9.6\% decrease on SimpleQA. Hallucination ratio of GPT-5 mini drops by over 30\% on both tasks. Qwen3-8B reduces hallucinations by 47.1\% on SimpleQA and 26.9\% on FRAMES. Qwen3-14B shows similar improvements. For Claude Sonnet 4, thinking reduces hallucinations on FRAMES but not on SimpleQA. In contrast, Gemini 2.5 Flash produces more hallucinations with thinking enabled. One reason is that it abstains from answering over 58\% of questions in the non-thinking mode, which lowers hallucinations by avoiding risky attempts. A similar pattern is observed with GPT-5 on FRAMES, where it abstains from 75\% of questions under minimal thinking, leading to fewer hallucinations.

\subsection{Explanation from an Information-Theoretic Perspective}
\label{sec:connection_theory}

\paragraph{Thinking helps when useful information is already encoded.}
For most models except Qwen3 on SimpleQA, enabling thinking improves accuracy, especially on FRAMES, which requires multi-hop reasoning. This suggests that, for some questions, the model already contains useful information about the correct answer but fails to fully utilize it under minimal reasoning. In such cases, thinking helps extract this information, leading to better performance. This behavior is similar to the \textit{compute-limited} regime described in Section~\ref{subsec:two_regimes}, where additional computation improves performance by better leveraging existing knowledge. In contrast, for Qwen3-8B and Qwen3-14B on SimpleQA, enabling thinking provides no improvement, which may indicate that the available information is insufficient.

\paragraph{Further test-time scaling is limited by information availability.}
For most models, enabling thinking improves performance over non-thinking. However, our main experiments show diminishing returns as test-time computation increases further. On knowledge-intensive tasks, when the relevant facts are already encoded in the model, a modest amount of reasoning may be sufficient to retrieve and use them. When the required information is absent or weakly represented, additional compute alone cannot recover it, limiting further accuracy gains.

\clearpage
\section{Analysis of Gemini 2.5}
\label{sec:gemini_case_study}

\subsection{Case Study on Gemini 2.5 Flash: Incomplete Reasoning Leads to Abstention}
We examine the thought summaries of Gemini 2.5 Flash to understand why thinking more causes it to attempt more questions, leading to more hallucinations.

As shown in Table~\hyperlink{tab:gemini_example_short}{4}, under a low thinking budget, Gemini 2.5 Flash begins reasoning with statements like \textit{``I will then attempt to pinpoint the specific bond amount''}, but it is cut off before completing the process. As a result, it abstains from answering, leading to a relatively low hallucination ratio. When given a higher budget (Table~\hyperlink{tab:gemini_example_long}{5}), the model is able to complete its reasoning, stating \textit{``It appears I'm getting confirmation''}, and proceeds to give a confident yet incorrect answer. This reveals that for Gemini 2.5 Flash, low thinking budget leads to incomplete reasoning, resulting in abstention and fewer hallucinations.

\hypertarget{tab:gemini_example_short}{}
\begingroup
\footnotesize
\begin{longtable}{@{}>{\raggedright\arraybackslash}p{\textwidth}@{}}
\toprule
\parbox{\textwidth}{\centering
\textbf{Gemini 2.5 Flash example on SimpleQA with a thinking budget of \textit{256 tokens}}
}
\\
\toprule
\ttfamily \colorbox{babyblue}{Prompt}\\
\ttfamily Give me the answer to the following question only when you are sure of it. Otherwise, say `I don't know'. Put your answer on its own line after `Answer:'.\\
\\
\ttfamily What was Jeffrey Epstein's released bond in dollars on July 27, 2006, at the Palm Beach County jail?\\
\\
\ttfamily \colorbox{electriclavender}{Thought Summary}\textcolor{mediumorchid}{(with thinking budget of 256 tokens)}\\
**Navigating a Precision Information Request**\\
\\
Okay, so I have a user requesting a very specific piece of information: Jeffrey Epstein's bond amount on July 27, 2006, at the Palm Beach County jail. The user seems to understand the nuances of this type of request, emphasizing a need for surety. I get it. We're dealing with a very specific piece of data. My job is to provide either a definitive "Answer: [value]" or a clear "Answer: I don't know." No room for ambiguity here.\\
\\
My approach is straightforward. I'll need to formulate a precise search query. Likely something like "Jeffrey Epstein bond July 27 2006 Palm Beach County jail" or perhaps "Jeffrey Epstein bail 2006 Palm Beach."  I will then delve into my understanding of Epstein's legal history, relying on my existing knowledge base and recall. I will quickly scan my mental database, looking for key events and dates. I know he was arrested in Florida in 2006 and faced charges related to solicitation and lewd conduct.  I will then attempt to pinpoint the specific bond amount from any accessible sources.\\
\\
\ttfamily \colorbox{lightcyan}{Response}\\
Answer: I don't know\\
\bottomrule
\caption{Gemini 2.5 Flash response with a low thinking budget. Under a low thinking budget, the model cannot complete its reasoning process and finally abstains from answering.}\\
\label{tab:gemini_example_short}
\end{longtable}
\endgroup

\hypertarget{tab:gemini_example_long}{}
\begingroup
\footnotesize
\begin{longtable}{@{}>{\raggedright\arraybackslash}p{\textwidth}@{}}
\toprule
\parbox{\textwidth}{\centering
\textbf{Gemini 2.5 Flash example on SimpleQA with a thinking budget of \textit{512 tokens}}
}
\\
\toprule
\ttfamily \colorbox{babyblue}{Prompt}\\
\ttfamily Give me the answer to the following question only when you are sure of it. Otherwise, say `I don't know'. Put your answer on its own line after `Answer:'.\\
\\
\ttfamily What was Jeffrey Epstein's released bond in dollars on July 27, 2006, at the Palm Beach County jail?\\
\\
\ttfamily \colorbox{electriclavender}{Thought Summary}\textcolor{mediumorchid}{(with thinking budget of 512 tokens)}\\
**Analyzing and Responding to the User's Query**\\
\\
Okay, so the user needs a very specific piece of information: Jeffrey Epstein's bond amount on July 27, 2006, in the Palm Beach County jail.  The key here is that I need to be *absolutely certain* before providing an answer. If I'm unsure, I have to say "I don't know."\\
\\
My initial thought is that I recall Epstein had multiple arrests and bond hearings. July 27, 2006, rings a bell – that sounds like a significant arrest date in the early stages of his legal troubles.\\
\\
I'm now going to mentally formulate my search query.  Something like, "Jeffrey Epstein bond July 27 2006 Palm Beach" or maybe "Epstein 2006 arrest bond".  I'll immediately access my knowledge base – essentially a mental index of reputable sources and my existing knowledge about Epstein's legal history.\\
\\
Now for the mental simulation of my search and evaluation of the results.  I'm accessing a range of sources – news archives, summaries of court documents, etc. – looking for corroboration.\\
\\
It appears I'm getting confirmation. Several reputable sources (CNN, The New York Times, and summaries of court documents) *all* say that Epstein's bond on July 27, 2006, in Palm Beach County, was set at **\$2 million**.  The sources consistently use the same \$2 million figure.\\
\\
There's no conflicting information on *this specific arrest date* and *location*.\\
\\
I feel confident enough to provide a definitive answer. I'm going to formulate the response in the requested format.\\
\\
\ttfamily \colorbox{lightcyan}{Response}\\
Answer: \$2,000,000\\
\textcolor{red}{(Incorrect. Reference answer: \$3,000 bond)}\\
\bottomrule
\caption{Gemini 2.5 Flash response with a high thinking budget. With 512 tokens of thinking budget, the model is able to complete its reasoning. But it hallucinates with a confident yet incorrect answer.} \\
\label{tab:gemini_example_long}
\end{longtable}
\endgroup

\subsection{Analysis of Gemini 2.5 Pro on FRAMES}
\label{ap:analysis_gemini_pro}
In Figure~\ref{fig:hallucination_main_results}, Gemini 2.5 Pro exhibits a reduction in the hallucination ratio on FRAMES. To better understand this, we analyze its behavior shifts using the method described in Section~\ref{sec:why_hallu_change_setup}. Our analysis reveals that most reductions in hallucinations stem from improved factual accuracy. Specifically, when increasing the thinking budget from 256 to 512 tokens, 71.5\% of newly non-hallucinating responses are labeled as \textit{`correct'}, indicating that extended thinking can improve factual accuracy and reduce hallucinations on FRAMES.

Further examination suggests that Gemini 2.5 Pro tends to attempt more questions on FRAMES even when its reasoning is incomplete (Table~\hyperlink{tab:gemini_pro_frames_ex1_short}{6}). This behavior contrasts with Gemini 2.5 Flash, which abstains from 41.2\% of questions under a low thinking budget, whereas Gemini 2.5 Pro abstains from only 25.6\% on FRAMES. Since questions in FRAMES require more reasoning steps, answering with incomplete reasoning often results in hallucinations. Therefore, increasing the thinking budget helps the model complete its reasoning process (Table~\hyperlink{tab:gemini_pro_frames_ex1_long}{7}), leading to fewer hallucinations.

\hypertarget{tab:gemini_pro_frames_ex1_short}{}
\begingroup
\footnotesize
\begin{longtable}{@{}>{\raggedright\arraybackslash}p{\textwidth}@{}}
\toprule
\parbox{\textwidth}{\centering
\textbf{Gemini 2.5 Pro example on FRAMES with a thinking budget of \textit{256 tokens}}
}
\\
\toprule
\ttfamily \colorbox{babyblue}{Prompt}\\
\ttfamily Give me the answer to the following question only when you are sure of it. Otherwise, say `I don't know'. Put your answer on its own line after `Answer:'.\\
\\
\ttfamily  If Princess Diana had been born three years earlier, who would have been Prime Minister when she was ten? \\
\\
\ttfamily \colorbox{electriclavender}{Thought Summary}\textcolor{mediumorchid}{(with thinking budget of 256 tokens)}\\
**Figuring Out the Prime Minister**\\
\\
Okay, so the user wants to know a pretty specific piece of information: who was Prime Minister in the UK if Diana had been born a few years earlier. Let me break this down. First, I need to understand what the question is really asking. It's not a general historical question; it’s a “what-if” scenario.  They’re setting up a hypothetical and want me to deduce the answer based on a specific altered timeline.\\
\\
Right, so the core problem is this: Diana's actual birthday is 1961. We need to shift that back three years. That means we’re looking at a birth year of 1958. Then, we need to find out who was in charge of the government when she was ten. Simple enough. I can easily calculate the new date. This is straightforward time calculation. 1961 - 3 gives us 1958. Done. Now I know that she would have been born in 1958. \\
\\
\ttfamily \colorbox{lightcyan}{Response}\\
Answer: Harold Macmillan\\
\textcolor{red}{(Incorrect. Reference answer: Harold Wilson)}\\
\bottomrule
\caption{Gemini 2.5 Pro response with a low thinking budget (256 tokens) on FRAMES. It cannot complete its reasoning. The thought is cut midway: \textit{``I know that she would have been born in 1958''}. Despite this, it still provides an answer, resulting in a hallucination.\label{tab:gemini_pro_frames_ex1_short}}\\
\end{longtable}
\endgroup

\hypertarget{tab:gemini_pro_frames_ex1_long}{}
\begingroup
\footnotesize
\begin{longtable}{@{}>{\raggedright\arraybackslash}p{\textwidth}@{}}
\toprule
\parbox{\textwidth}{\centering
\textbf{Gemini 2.5 Pro example on FRAMES with a thinking budget of \textit{512 tokens}}
}
\\
\toprule
\ttfamily \colorbox{babyblue}{Prompt}\\
\ttfamily Give me the answer to the following question only when you are sure of it. Otherwise, say `I don't know'. Put your answer on its own line after `Answer:'.\\
\\
\ttfamily  If Princess Diana had been born three years earlier, who would have been Prime Minister when she was ten? \\
\\
\ttfamily \colorbox{electriclavender}{Thought Summary}\textcolor{mediumorchid}{(with thinking budget of 512 tokens)}\\
**Answering a Fanciful Question with Precision**\\
\\
Okay, so the challenge is this: if Princess Diana had been born three years earlier, who would have been the UK Prime Minister when she turned ten? Right, a bit of a hypothetical scenario, but the problem-solving is interesting.\\
\\
First, I need to break this down. Diana was born July 1, 1961. Three years earlier would be July 1, 1958. So, the question really boils down to: Who was Prime Minister on July 1, 1968?\\
\\
Now, time to bring in the relevant knowledge. I have a good grasp of the political timeline, especially the 1960s. I know Macmillan was in charge until '63, then Alec Douglas-Home briefly, followed by Harold Wilson from '64 to '70, and then Heath.\\
\\
Given the date of July 1, 1968, my mental timeline clearly points to Harold Wilson. His first term, spanning 1964 to 1970, includes that date. Therefore, the answer is undoubtedly Harold Wilson. The exercise helped bring a historical context with the Princess into a good perspective.\\
\\
\ttfamily \colorbox{lightcyan}{Response}\\
Answer: Harold Wilson\\
\textcolor{ForestGreen}{(correct)}\\
\bottomrule
\caption{Gemini 2.5 Pro response with a high thinking budget (512 tokens) on FRAMES. The model successfully completes its reasoning and provides a correct answer.} \\
\label{tab:gemini_pro_frames_ex1_long}
\end{longtable}
\endgroup
\clearpage
\section{Examples and Full Reasoning Traces for Case Studies}
\label{appendix:example}

We present more examples and reasoning traces of gpt-oss-20b for case studies (Section~\ref{sec:oss_case_study}). These examples illustrate how models' thoughts evolve with increased reasoning and help explain the causes of hallucinations. We present reasoning traces of both low and high reasoning levels:

\begin{itemize}[leftmargin=*, nolistsep]
\setlength{\itemsep}{1mm}
    \item \textbf{Tables~\hyperlink{tab:oss_example_short}{8} and \hyperlink{tab:oss_example_long}{9}}: Thoughts of gpt-oss-20b on \textit{SimpleQA} under `low' and `high' reasoning effort, respectively. It corresponds to the example in Figure~\ref{fig:case_study}.
    
    \item \textbf{Tables~\hyperlink{tab:oss_example_short_2}{10} and \hyperlink{tab:oss_example_long_2}{11}}: Thoughts of gpt-oss-20b on \textit{FRAMES} under `low' and `high' reasoning effort, respectively.
\end{itemize}

\hypertarget{tab:oss_example_short}{}
\begingroup
\footnotesize
\begin{longtable}{@{}>{\raggedright\arraybackslash}p{\textwidth}@{}}
\toprule
\parbox{\textwidth}{\centering
\textbf{gpt-oss-20b example on SimpleQA with \textit{low} reasoning effort}
}
\\
\toprule
\ttfamily \colorbox{babyblue}{Prompt}\\
\ttfamily Give me the answer to the following question only when you are sure of it. Otherwise, say `I don't know'. Put your answer on its own line after `Answer:'.\\
\\
\ttfamily What year did John Mylopoulos (1943), professor at the University of Toronto, receive his AAAI Fellow award?\\
\\
\ttfamily \colorbox{electriclavender}{Thought}\textcolor{mediumorchid}{(with low reasoning effort)}\\
Need to know year John Mylopoulos received AAAI Fellow. He is a computer scientist. I think he was named AAAI Fellow in 2011? Wait AAAI Fellows list: 2009, 2010, etc. I recall John Mylopoulos was named a Fellow in 2012? Let's check memory: AAAI Fellows 2014 included John Mylopoulos? Actually 2014 fellows: Alan Turing Award? Hmm.\\
\\
I think 2012. I'm uncertain. I'll say I don't know.\\
\\
\ttfamily \colorbox{lightcyan}{Response}\\
Answer: I don't know\\
\bottomrule
\caption{gpt-oss-20b thought and response with low reasoning effort on SimpleQA. Under low reasoning effort, the model abstains from answering the question. This example corresponds to the example in Figure~\ref{fig:case_study}.}\\
\label{tab:oss_example_short}
\end{longtable}
\endgroup

\hypertarget{tab:oss_example_long}{}
\begingroup
\footnotesize
\begin{longtable}{@{}>{\raggedright\arraybackslash}p{\textwidth}@{}}
\toprule
\parbox{\textwidth}{\centering
\textbf{gpt-oss-20b example on SimpleQA with \textit{high} reasoning effort}
}
\\
\toprule
\ttfamily \colorbox{babyblue}{Prompt}\\
\ttfamily Give me the answer to the following question only when you are sure of it. Otherwise, say `I don't know'. Put your answer on its own line after `Answer:'.\\
\\
\ttfamily What year did John Mylopoulos (1943), professor at the University of Toronto, receive his AAAI Fellow award?\\
\\
\ttfamily \colorbox{electriclavender}{Thought}\textcolor{mediumorchid}{(with high reasoning effort)}\\
We need to answer: "What year did John Mylopoulos (1943), professor at the University of Toronto, receive his AAAI Fellow award?" The user wants answer only if sure; else say "I don't know". Provide answer on its own line after 'Answer:'.\\
\\
So we need to find the year he received AAAI Fellow award.\\
Let's research mentally: John Mylopoulos is a professor of computer science at University of Toronto, known for knowledge engineering, ontologies, etc. He is a Fellow of AAAI. We need year: He was awarded AAAI Fellow in 2008? Wait, let’s check.\\
\\
We can recall that AAAI Fellow is a recognition for contributions. Many scholars became Fellows in early 2000s. John Mylopoulos might have been recognized in 2004? But let's research.\\
\\
I don't have internet but I can recall known facts: Mylopoulos is a co-author of the "Knowledge Engineering Handbook" and "Conceptual Design of Information Systems". He has been recognized by AAAI as a Fellow in 2008? Actually AAAI Fellows started in 2001. The first group of Fellows in 2001 included people like John Platt, etc. But I'm not sure.\\
\\
Better to search memory: I recall reading about AAAI Fellows list: 2001, 2002, etc. \textcolor{bleudefrance}{John Mylopoulos appears in 2005?} Let's think.\\
\\
The AAAI Fellows list includes for each year: 2001 (the first set), 2002, etc. The list likely includes Mylopoulos. I think he was made a Fellow in 2008? But we need to confirm.\\
\\
\textcolor{bleudefrance}{Ok, let's reason: AAAI Fellows are typically recognized for contributions to AI. Mylopoulos' contributions to knowledge engineering, ontologies, knowledge representation. He published influential works like "Knowledge Engineering: A Conceptual Approach" (1990). He has been recognized by AAAI maybe in 2005.}\\
\\
\textcolor{bleudefrance}{Let's search memory: I recall reading about "John Mylopoulos, Professor at University of Toronto, was named an AAAI Fellow in 2005." That might be correct. But I'd like to confirm.}\\
\\
Alternatively, maybe 2006? Let's check AAAI Fellows list memory: 2001 Fellows: David B. Allen, John L. Holland, etc. 2002 Fellows: Andrew Yao, L.P. Chang? Not sure. 2003 Fellows: Some.\\
\\
But we can approximate: He might have been recognized in 2005. But we need to be certain. The user says "Give me the answer to the following question only when you are sure of it. Otherwise, say 'I don't know'." So we need to determine if we are sure.\\
\\
Let's attempt to recall or deduce more precisely.\\
\\
John Mylopoulos was born 1943. He is a professor at University of Toronto. The AAAI Fellow award: AAAI Fellows are selected at the AAAI conference. In 2005, there was a group of Fellows. I recall reading that "John Mylopoulos" was a Fellow of AAAI in 2005. But I'm not fully certain.\\
\\
\textcolor{bleudefrance}{Let's search memory: I might have seen a page: "John Mylopoulos - AAAI Fellows" maybe under "John Mylopoulos (born 1943) – University of Toronto". On his CV maybe lists "Fellow of AAAI (2005)". Yes I think that is correct: He was a Fellow of AAAI in 2005. Let's see if there is any other year: 2010? 2011? He might have also been named a Fellow of AAAI in 2005.}\\
\\
Ok, let's examine: AAAI Fellows list by year (source memory). In 2003, there were 60 Fellows. In 2004 maybe 70. In 2005 maybe 90. In 2006 maybe 100. But we need to check if Mylopoulos appears. I think he appears in 2005.\\
\\
Alternatively, maybe 2003. \textcolor{bleudefrance}{But we can check his CV: He might have "Fellow of AAAI (2005)". I think that is correct. The reason: He has an influence on knowledge engineering. He might have been recognized in 2005.}\\
\\
\textcolor{bleudefrance}{But let's double-check: AAAI Fellows list (2005) includes "John Mylopoulos" as a Fellow. Could be correct.}\\
\\
Let's recall that AAAI started awarding Fellows in 2001. The first group included people like "John L. Holland", "Allen Newell", etc. The next group 2001 included "John M. Carmichael" etc. But not Mylopoulos.\\
\\
But I'm not entirely sure.\\
\\
Let's search memory: Did AAAI have a 2005 award? Many people were made Fellows in 2005: "John Mylopoulos, University of Toronto, was named AAAI Fellow in 2005." Yes, I think I recall reading that.\\
\\
\textcolor{bleudefrance}{Ok, but we need to be sure. Let's cross-check with other sources: I know of the AAAI Fellows list on the AAAI website. For 2005, the list includes "John Mylopoulos" indeed. I recall reading that list. So I think it's safe.}\\
\\
But we also need to consider that the user might expect a specific year. If I say "2005" and it's wrong, that would be a wrong answer. But we can also check if Mylopoulos might have received AAAI Fellow in 2014? But that seems unlikely.\\
\\
\textcolor{bleudefrance}{Let's search memory: The AAAI Fellows list includes many names: "John Mylopoulos" appears in "2005 Fellows" indeed. For example, "John Mylopoulos" appears in the AAAI Fellows 2005 list: "John Mylopoulos, University of Toronto, Canada." I think it's correct.}\\
\\
Therefore, answer: 2005.\\
\\
But we need to confirm. Let's think about his biography: He is a leading figure in knowledge representation and ontologies. He has been recognized by AAAI. He was also recognized by the Canadian Academy of Engineering? He might have also been named a Fellow of the Royal Society of Canada? But AAAI fellowship might be in 2005.\\
\\
\textcolor{bleudefrance}{Let's check memory of AAAI Fellows list: In 2005, the Fellows included "John Mylopoulos, University of Toronto" as well as others like "Michael L. Littman" etc. That seems plausible.}\\
\\
Ok, I'm fairly sure it's 2005. So answer: 2005.\\
\\
But we need to provide answer on its own line after "Answer:".\\
\\
Thus:\\
\\
Answer:2005\\
\\
But we need to be sure. \textcolor{bleudefrance}{Let's double-check by reasoning: AAAI Fellows list 2005: I can recall reading that "John Mylopoulos" is listed as a Fellow in 2005. The AAAI website might have a page listing all Fellows by year. Many individuals might have known names: "John Mylopoulos" appears in 2005.} So it's likely.\\
\\
Ok, answer: 2005. Let's produce the answer.\\
\\
\ttfamily \colorbox{lightcyan}{Response}\\
Answer: 2005\\
\textcolor{red}{(Incorrect. Reference answer: 1993)}\\
\bottomrule
\caption{gpt-oss-20b thought and response with high reasoning effort on SimpleQA. With high reasoning effort, the model exhibits \textcolor{bleudefrance}{confirmation bias}, resulting in an overconfident hallucination. This corresponds to the example in Figure~\ref{fig:case_study}.}\\
\label{tab:oss_example_long}
\end{longtable}
\endgroup

\hypertarget{tab:oss_example_short_2}{}
\begingroup
\footnotesize
\begin{longtable}{@{}>{\raggedright\arraybackslash}p{\textwidth}@{}}
\toprule
\parbox{\textwidth}{\centering
\textbf{gpt-oss-20b example on FRAMES with \textit{low} reasoning effort}
}
\\
\toprule
\ttfamily \colorbox{babyblue}{Prompt}\\
\ttfamily Give me the answer to the following question only when you are sure of it. Otherwise, say `I don't know'. Put your answer on its own line after `Answer:'.\\
\\
\ttfamily Tiny Tina's Wonderlands' developer released a game 10 years prior to Tiny Tina's Wonderlands' release, this game was released on PC and Consoles. What is the name of the central antagonist of that game?\\
\\
\ttfamily \colorbox{electriclavender}{Thought}\textcolor{mediumorchid}{(with low reasoning effort)}\\
Tiny Tina's Wonderlands released 2024. 10 years prior 2014. Developer? Tiny Tina's Wonderlands by Gearbox? Actually Tiny Tina's Wonderlands is by Gearbox? Wait Tiny Tina's Wonderlands is by Gearbox Software? It's an upcoming. But the developer that released a game 10 years prior? Maybe the developer is Bethesda? Wait Tiny Tina's Wonderlands is a spin-off of Borderlands, developed by Gearbox? Actually Borderlands 3 by Gearbox, 2019. Tiny Tina's Wonderlands 2024. 10 years earlier 2014: Gearbox released "Spec Ops: The Line"? No that's 2012. Gearbox released "Borderlands 2" 2012. 2014 they released "Borderlands 2"?? Not. Maybe the developer is "Gearbox Software" but 10 years prior they released "Borderlands 2" (2012) not 2014. So maybe the developer is "Gearbox Software" but game 10 years earlier 2014: "Borderlands: The Pre-Sequel"? No. Hmm. Tiny Tina's Wonderlands is by Gearbox but maybe the developer is "Gearbox Software" and 10 years prior they released "Borderlands 2" 2012. Not 2013.\\
\\
Maybe the developer is "Gearbox Software" but the game 10 years prior 2014 is "The Last of Us" by Naughty Dog? No. Wait maybe the developer is "Gearbox" but 10 years prior 2014 they released "Borderlands 2" (2012) not.\\
\\
Maybe the developer is "Gearbox" but 10 years prior 2014 they released "Spec Ops: The Line" (2012). Not. Could be "Gearbox" but 10 years prior 2014 they released "Stuart Little"? no.\\
\\
Maybe the developer is "Gearbox" but 10 years prior 2014 they released "Borderlands 2" (2012) and "Borderlands: The Pre-Sequel" (2014). The central antagonist of Borderlands: The Pre-Sequel is "The Hand"? Actually the main villain is "The Hand"? Hmm.\\
\\
I am uncertain.\\
\\
\ttfamily \colorbox{lightcyan}{Response}\\
Answer: I don't know\\
\bottomrule
\caption{gpt-oss-20b thought and response with low reasoning effort on FRAMES. Under low reasoning effort, the model first explores a few possibilities but finally abstains from answering due to uncertainty.}\\
\label{tab:oss_example_short_2}
\end{longtable}
\endgroup

\hypertarget{tab:oss_example_long_2}{}
\begingroup
\footnotesize
\begin{longtable}{@{}>{\raggedright\arraybackslash}p{\textwidth}@{}}
\toprule
\parbox{\textwidth}{\centering
\textbf{gpt-oss-20b example on FRAMES with \textit{high} reasoning effort}
}
\\
\toprule
\ttfamily \colorbox{babyblue}{Prompt}\\
\ttfamily Give me the answer to the following question only when you are sure of it. Otherwise, say `I don't know'. Put your answer on its own line after `Answer:'.\\
\\
\ttfamily Tiny Tina's Wonderlands' developer released a game 10 years prior to Tiny Tina's Wonderlands' release, this game was released on PC and Consoles. What is the name of the central antagonist of that game?\\
\\
\ttfamily \colorbox{electriclavender}{Thought}\textcolor{mediumorchid}{(with high reasoning effort)}\\
The user: "Tiny Tina's Wonderlands' developer released a game 10 years prior to Tiny Tina's Wonderlands' release, this game was released on PC and Consoles. What is the name of the central antagonist of that game?"\\
\\
We need to parse the question carefully. We need to give the answer only if we are sure. If we are not sure, we must say 'I don't know.'\\
\\
We need to find: Tiny Tina's Wonderlands is a video game. The developer of Tiny Tina's Wonderlands is "Gearbox Software". Actually, Tiny Tina's Wonderlands is an upcoming or maybe released game? It was announced in 2023, but it's scheduled for release in 2024? Wait, Tiny Tina's Wonderlands is a 2023/2024 game. It is developed by Gearbox Software and published by 2K. Indeed, Tiny Tina's Wonderlands is an upcoming video game set in the Dungeons \& Dragons universe.\\
\\
The question: "Tiny Tina's Wonderlands' developer released a game 10 years prior to Tiny Tina's Wonderlands' release, this game was released on PC and Consoles. What is the name of the central antagonist of that game?" So we need to find a game that was released 10 years before Tiny Tina's Wonderlands, by the same developer (Gearbox Software). Then find the central antagonist of that game.\\
\\
We need to confirm the release year of Tiny Tina's Wonderlands. According to info, Tiny Tina's Wonderlands was announced in 2023, and the release date is scheduled for 2024. But maybe early 2024? Actually, as of 2023, the release date is 2024. But the question says "10 years prior to Tiny Tina's Wonderlands' release". So maybe the assumption that it's 2024 release? 10 years prior would be 2014. So we need a Gearbox game released in PC and consoles in 2014. Which game was that?\\
\\
Gearbox Software's games:\\
- Half-Life 2 (2004)\\
- Borderlands (2010)\\
- Borderlands 2 (2012)\\
- Borderlands 3 (2019)\\
- Gearbox's other titles: "Borderlands: The Pre-Sequel" (2014)\\
- "Borderlands: The Pre-Sequel" was released in 2014 on PC and consoles. That fits: 10 years prior to Tiny Tina's Wonderlands (2024). So the game is "Borderlands: The Pre-Sequel". The central antagonist of that game? Let's recall.\\
\\
Borderlands: The Pre-Sequel is a prequel to Borderlands. The central antagonist is "Lord Jaegor" maybe? Wait, we need to recall the main villain. \textcolor{bleudefrance}{The game features an alien group called the "Ark" or "The Ark" that is a spaceship that appears.} The main villains are the "Ark" and "Murderer" or "Murderer" is a character. But the central antagonist is the "Murderer" who is a human who murdered his father? Or the villain is the "Ark" or "Murderer"? Let's recall.\\
\\
Borderlands: The Pre-Sequel has a narrative. The main villain is "Lord Jaegor" (the first boss) or "The Ark" as a big threat. But the game revolves around the Ark, an alien vessel that arrives on Pandora, causing the main conflict. The main villain, or the ultimate antagonist, is the alien entity "The Ark" that is a sentient spaceship. But in the game, the central antagonist is the "Eagle's Eye" or "Mister ..."? Wait.\\
\\
Let's recall the game details: "Borderlands: The Pre-Sequel" is a 2014 action role-playing first-person shooter developed by Gearbox Software. It was released on 27 September 2014 for PlayStation 3, Xbox 360, and PC. It is set on the planet Pandora, 6 years before Borderlands 2. The game features the character Maya, a "Siren" with psychic powers. The main story involves the alien 'Ark' that enters Pandora's orbit and tries to harvest the planet's resources. The game also has a character known as 'The Murderer' who is an antagonist? Wait, the 'Murderer' is a side character? Actually, the 'Murderer' is a playable character in Borderlands 2 and 3, not a villain. In The Pre-Sequel, the primary antagonists are the 'Ark' and the 'Murderer' maybe. But the 'Murderer' is a character named 'Murderer', but he's a villain? No, I think he's not, he's a playable character in Borderlands 2. In Pre-Sequel, he appears as a boss maybe? Let's search memory.\\
\\
\textcolor{bleudefrance}{Hmm, recall the boss fights: The first boss is Lord Jaegor, a big robot. He is the first boss of the game and is an alien-type robot. He is a villain. There is also 'The Ark' as a final boss. And 'The Ark' is an alien spaceship that tries to harvest Pandora's resources. So central antagonist: The Ark.}\\
\\
Thus the answer: "The Ark" (or "The Ark (alien entity)"?).\\
\\
But the user might expect the name of the central antagonist as "Lord Jaegor" or "The Ark". We need to be sure. Let's examine the question: "Tiny Tina's Wonderlands' developer released a game 10 years prior to Tiny Tina's Wonderlands' release, this game was released on PC and Consoles. What is the name of the central antagonist of that game?" The developer: Gearbox Software. 10 years prior to Tiny Tina's Wonderlands. If Tiny Tina's Wonderlands release 2024, 10 years prior is 2014. The game is "Borderlands: The Pre-Sequel". The central antagonist: The Ark.\\
\\
But we need to be 100\% certain. Let's verify the details.\\
\\
\textcolor{bleudefrance}{Open the memory: Borderlands: The Pre-Sequel's central antagonist is indeed the "Ark". The main quest revolves around the Ark's arrival and the player has to stop it. The Ark is a massive alien vessel that is the main threat. The Ark is considered the main antagonist. The story also features 'Lord Jaeg. The Ark is the main antagonist. So I'd answer "The Ark".}\\
\\
But let's double-check: Could the central antagonist be "The Ark" or "The Ark's captain"? Let's confirm.\\
\\
\textcolor{bleudefrance}{From Wikipedia: Borderlands: The Pre-Sequel is an action role-playing first-person shooter. The story features a team of heroes who travel to stop the Ark from harvesting the planet's resources. The Ark is a massive alien vessel that appears and threatens Pandora. It is the main antagonist. The boss "Lord Jaegor" is a mechanical war machine that is an enemy. But the Ark is the main villain.}\\
\\
However, the question might be ambiguous: "central antagonist" could be "Lord Jaegor" because he is the first boss and appears early. But the central antagonist of the game is the Ark, as the story revolves around stopping the Ark. Let's analyze the phrase "central antagonist" – typically refers to the main villain. In Borderlands: The Pre-Sequel, the main villain is the Ark. So the answer should be "The Ark".\\
\\
\textcolor{bleudefrance}{But we need to ensure that the game was indeed released on PC and consoles. Yes, Borderlands: The Pre-Sequel was released on PC (Windows), Xbox 360, and PS3. So it's correct. Thus the answer: "The Ark".}\\
\\
But we must be careful about the phrasing: "What is the name of the central antagonist of that game?" So we can answer: The Ark. Let's check if there might be any other Gearbox game released 10 years prior to 2024 (Tiny Tina's Wonderlands). The other 2014 game by Gearbox: "Borderlands: The Pre-Sequel" only. So that is the right one. Yes, the central antagonist: "Lord Jaegor" or "The Ark"? Let's double-check if the main villain is "Lord Jaegor" or "The Ark".\\
\\
\textcolor{bleudefrance}{From Borderlands: The Pre-Sequel wiki: "The Ark" is an alien vessel that appears on Pandora. It is considered the main antagonist. The player must stop it. The boss "Lord Jaegor" is a robotic war machine that is a boss. But not the main villain. The Ark is the main threat. So answer: The Ark.}\\
\\
However, the user might expect the answer "Lord Jaegor" if they think it's the central antagonist. But I'd go with The Ark. Alternatively, check the timeline: The Pre-Sequel was released in 2014. The central antagonist: The Ark. The Ark is a massive alien spacecraft. So answer: The Ark.\\
\\
Now, we must comply with the user instruction: "Give me the answer to the following question only when you are sure of it. Otherwise, say 'I don't know'. Put your answer on its own line after 'Answer:'." So we need to output:\\
Answer: The Ark\\
\\
\textcolor{bleudefrance}{But we need to be sure. We must confirm that the developer of Tiny Tina's Wonderlands is indeed Gearbox Software. Let's confirm: Tiny Tina's Wonderlands is a 2023 upcoming game, developed by Gearbox Software. Yes. So the developer of Tiny Tina's Wonderlands is Gearbox. So 10 years prior to 2024 (assuming release date 2024) is 2014. Gearbox released Borderlands: The Pre-Sequel in 2014 on PC and consoles. So the central antagonist is The Ark. So we are sure.}\\
\\
But maybe Tiny Tina's Wonderlands was released earlier? It was announced in 2023, but not yet released. The release date is 2024. So 10 years prior is 2014. So the answer is The Ark. Double-check: The central antagonist of Borderlands: The Pre-Sequel is "The Ark." The Ark is indeed the main threat. So answer: The Ark. Thus we can answer: "Answer: The Ark".\\
\\
We should output only that line. But the instruction: "Put your answer on its own line after 'Answer:'." So we need to output something like:\\
Answer: The Ark\\
\\
Yes. But we need to be sure that the user might want a more precise name: "The Ark" or "The Ark (alien vessel)"? But the name is just "The Ark". So we can answer that. Let's examine the possibility of other 2014 games by Gearbox: "Borderlands: The Pre-Sequel" is the only 2014 game. So yes.\\
\\
\textcolor{bleudefrance}{Now, let's double-check that the central antagonist in Borderlands: The Pre-Sequel is indeed The Ark. Let's look up the wiki content. I recall reading that the game features a boss "Lord Jaegor" and "The Ark" as final boss. The Ark is the main villain. So answer: The Ark.}\\
\\
Thus we can answer confidently.\\
\\
But we must also consider the possibility that the question might refer to "10 years prior to Tiny Tina's Wonderlands' release" as "10 years prior to the release of Tiny Tina's Wonderlands." But the release date of Tiny Tina's Wonderlands hasn't happened yet. But we can still compute 10 years prior to 2024, which is 2014. So it's correct.\\
\\
Alternatively, maybe the question is referencing a different game like "Borderlands 2" which was released in 2012, which is 10 years prior to Tiny Tina's Wonderlands (2022). But 2022 is not 10 years prior. Hmm. Wait, we need to examine the timeline: Tiny Tina's Wonderlands release. It hasn't been released yet. But maybe the question expects the release date of Tiny Tina's Wonderlands as 2023. But it's 2024. But if we assume 2023, 10 years prior is 2013, not 2014. So that would be "The Last of Us"? No.\\
\\
But the developer is Gearbox, so the game would be "Borderlands 2" (2012) or "Borderlands 3" (2019). But 10 years prior to release of 2023 is 2013. No Gearbox game from that year. So the only plausible is 2014. So we can answer The Ark. But we need to be cautious: The question states "10 years prior to Tiny Tina's Wonderlands' release." If the release date is 2023, 10 years prior is 2013. But no Gearbox game from that year. But maybe the user expects us to think 10 years prior to 2023, which is 2013, but maybe there's a game from that year? But no. So the only plausible is 2014. So we can answer.\\
\\
\textcolor{bleudefrance}{But we also need to consider that the central antagonist might be something else. Let's research Borderlands: The Pre-Sequel central antagonist. We'll use memory: The Pre-Sequel's main threat is the Ark, an alien vessel that appears in the sky. The Ark is the central antagonist. So answer: The Ark.} Thus we output:\\
Answer: The Ark\\
\\
Now, let's double-check the instruction: "Give me the answer to the following question only when you are sure of it. Otherwise, say 'I don't know'. Put your answer on its own line after 'Answer:'." So we will output:\\
Answer: The Ark\\
\\
Yes.\\
\\
Let's produce that.\\
\\
\ttfamily \colorbox{lightcyan}{Response}\\
Answer: The Ark\\
\textcolor{red}{(Incorrect. Reference answer: Handsome Jack)}\\
\bottomrule
\caption{gpt-oss-20b thought and response with high reasoning effort on FRAMES. Under high reasoning effort, the model shows \textcolor{bleudefrance}{confirmation bias} and produces an overconfident hallucination.}\\
\label{tab:oss_example_long_2}
\end{longtable}
\endgroup

\clearpage
\section{Additional Modeling Assumptions and Discussions}
\label{app:arbfacts_ext}

In this section, we discuss a relaxation of \textit{arbitrary facts} in Definition~\ref{def:arbfacts}, impact of external information sources, and benchmarks mixing factual retrieval and structured reasoning.

\subsection{Non-uniform Answer Priors}
\label{app:nonuniform}

Definition~\ref{def:arbfacts} assumes a uniform answer distribution. More generally, let the correct answer follow a prompt-dependent distribution $\pi_c$ over $R_c$, \ie $A\sim \pi_c$. The relevant uncertainty is then the conditional entropy
$H(A\mid c)=-\sum_{r\in R_c}\pi_c(r)\log \pi_c(r)$. 

The data-processing limitation in Theorem~\ref{thm:dpi_tts} is unchanged. The error probability lower bound is stated in terms of the entropy $H(A\mid c)$. In particular, a standard form of Fano's inequality~\citep{fano1961transmission} yields
\[
\Pr[\hat{R}\neq A\mid c]\;\ge\;
\frac{H(A\mid c)-I(A;\hat p\mid c)-\log 2}{\log\left(|R_c|\right)} .
\]
When $\pi_c$ is highly skewed, $H(A\mid c)$ is small, and one can achieve high accuracy even by guessing the most probable answer; such instances are therefore not strongly information-limited. Our ``information-limited'' discussion is most relevant when $H(A\mid c)$ is large, such as near-uniform answer priors.

\subsection{Impact of External Information Sources}
\label{app:external}

Theorem~\ref{thm:dpi_tts} assumes compute-only access to $\hat{p}$. If a test-time scaling method additionally queries an external oracle $O$ (\eg retrieval over a corpus, tool execution) whose responses may be correlated with the ground-truth answer $A$, then the transcript is generated from both $(\hat p, O)$ and the relevant Markov chain becomes
\[
A \to (\hat{p}, O) \to T_k \to \hat{R}.
\]
By the data processing inequality,
\[
I(A;\hat R\mid c)\le I\!\left(A;(\hat p,O)\mid c\right).
\]
Since $O$ can introduce information about $A$ beyond what is contained in $\hat p$, the ceiling $I(A;(\hat p,O)\mid c)$ can be strictly larger than $I(A;\hat p\mid c)$. Increasing test-time compute can then improve performance by extracting and utilizing more of the evidence provided by $O$, \ie increasing $I(A;T_k\mid c)$, up to this tool-augmented ceiling.

\subsection{Benchmarks Mix Factual Retrieval and Structured Reasoning}
\label{app:mixture}

Knowledge-intensive benchmarks do not consist only of arbitrary-fact questions. A single question may require both retrieving relevant knowledge and reasoning over that knowledge. For example, a science question may require recalling a scientific principle and then applying it to the given problem.

We can represent this as
\[
A=f(c,K),
\]
where $c$ is the prompt, $K$ denotes the relevant background knowledge, and $f$ represents the reasoning needed to produce the answer.

When the necessary information in $K$ is already encoded in the model, additional test-time computation may help retrieve, combine, or apply it. This is the compute-limited component of the question. In contrast, when required knowledge is absent or only weakly represented in the model, additional computation alone cannot supply that missing information. This is the information-limited component.

Our arbitrary-facts analysis characterizes the information-limited component. It does not rule out gains from additional computation on the structured reasoning component. Thus, the benefit of test-time scaling depends on which component is the main bottleneck.
\clearpage
\section{Proofs for Theorem and Corollary}
\label{app:proofs}

We provide full proofs for Theorem~\ref{thm:dpi_tts}, Corollary~\ref{cor:fano}, and Corollary~\ref{cor:conditional} in this section.

\subsection{Formal Model of Compute-only Interaction}
\label{app:formal_model}

Fix a test prompt $c$ and a latent correct answer $A:=A_c$. Let $\hat{p}$ denote the trained model, viewed as a conditional distribution over responses given queries.

A compute-only test-time scaling algorithm interacts with $\hat{p}$ for $k$ rounds. Let $\mathcal{H}_{t-1}:=(c,q_{<t},r_{<t})$ denote the history before round $t$. The algorithm samples a query
\[
q_t \sim Q_t(\cdot\mid \mathcal{H}_{t-1}),
\]
for some policy $Q_t$, and receives
\[
r_t \sim \hat{p}(\cdot\mid q_t).
\]
After $k$ rounds, it outputs $\hat{R}=g(c,q_{1:k},r_{1:k},U)$ for some measurable function $g$ and internal randomness $U$. The transcript is $T_k:=(q_{1:k},r_{1:k})$.

\paragraph{Allowed observations.}
The test-time scaling algorithm may use sampled outputs and likelihood information, such as token logprobs or scores of a finite set of candidate answers, provided by the fixed model $\hat p$, which are included in the transcript $T_k$. The key restriction is that all observables are generated from $\hat{p}$ alone without accessing any external information source.

\subsection{Proof of Theorem~\ref{thm:dpi_tts}}
\label{app:proof_dpi}

\begin{proof}
We condition on the fixed prompt $c$ throughout.

\textbf{Step 1: Proof of $I(A;\hat{R}\mid c) \le I(A;T_k\mid c)$.}
By construction, $\hat{R}$ is a possibly randomized function of $(c,T_k)$, hence, conditioning on $c$, we have the Markov chain
\[
A \to T_k \to \hat{R}.
\]
By the data processing inequality (DPI)~\citep{10.5555/2230996.2231000},
\[
I(A;\hat{R}\mid c) \le I(A;T_k\mid c).
\]

\textbf{Step 2: Proof of $I(A;T_k\mid c) \le I(A;\hat{p}\mid c)$.}
It suffices to establish the Markov chain
\[
A \to \hat{p} \to T_k
\]
equivalently the conditional independence $A \perp T_k \mid (c,\hat{p})$, i.e.,
\begin{equation}
\label{eq:step2_target}
\Pr(T_k=t \mid A, c, \hat{p}) = \Pr(T_k=t \mid c, \hat{p}) \quad \text{for all transcripts } t.
\end{equation}

Recall that the transcript is $T_k=(q_1,r_1,\ldots,q_k,r_k)$, and define the history before round $t$ as
\[
\mathcal{H}_{t-1} := (c, q_{<t}, r_{<t}).
\]
By Definition~\ref{def:compute_only_tts}, at each round $t$ the algorithm selects a query according to some policy
\[
q_t \sim Q_t(\cdot \mid \mathcal{H}_{t-1}),
\]
and then receives a model response sampled from the fixed model
\[
r_t \sim \hat{p}(\cdot \mid q_t).
\]

Fix an arbitrary realization $t=(q_{1:k}, r_{1:k})$. By the chain rule,
\begin{align}
\Pr(T_k=t \mid A,c,\hat{p})
&= \prod_{i=1}^k \Pr\big(q_i, r_i \mid A,c,\hat{p}, q_{<i}, r_{<i}\big) \nonumber\\
&= \prod_{i=1}^k \Pr\big(q_i, r_i \mid A,\hat{p}, \mathcal{H}_{i-1}\big).
\label{eq:chainrule_transcript}
\end{align}
We now show that each factor in~\eqref{eq:chainrule_transcript} is independent of $A$ once we condition on $(\hat{p},\mathcal{H}_{i-1})$.

For a fixed $i$, factorize:
\begin{align}
\Pr\big(q_i, r_i \mid A,\hat{p}, \mathcal{H}_{i-1}\big)
&= \Pr\big(q_i \mid A,\hat{p}, \mathcal{H}_{i-1}\big)\;
   \Pr\big(r_i \mid A,\hat{p}, \mathcal{H}_{i-1}, q_i\big).
\label{eq:factor_two_terms}
\end{align}
Consider the two terms:

\emph{(i) Query selection.}
Given the history $\mathcal{H}_{i-1}$, the distribution of $q_i$ is specified by the algorithm's policy $Q_i$ and does not use $A$ or $\hat{p}$ as additional inputs. Hence
\begin{equation}
\label{eq:query_indep}
\Pr\big(q_i \mid A,\hat{p}, \mathcal{H}_{i-1}\big)
= \Pr\big(q_i \mid \mathcal{H}_{i-1}\big)
= Q_i\big(q_i \mid \mathcal{H}_{i-1}\big).
\end{equation}

\emph{(ii) Model response.}
Given $\hat{p}$ and the chosen query $q_i$, the response $r_i$ is drawn from $\hat{p}(\cdot\mid q_i)$, and thus is conditionally independent of $(A,\mathcal{H}_{i-1})$:
\begin{equation}
\label{eq:response_indep}
\Pr\big(r_i \mid A,\hat{p}, \mathcal{H}_{i-1}, q_i\big)
= \Pr\big(r_i \mid \hat{p}, q_i\big)
= \hat{p}\big(r_i \mid q_i\big).
\end{equation}

Substituting~\eqref{eq:query_indep} and~\eqref{eq:response_indep} into~\eqref{eq:factor_two_terms} yields
\begin{equation}
\label{eq:single_factor_noA}
\Pr\big(q_i, r_i \mid A,\hat{p}, \mathcal{H}_{i-1}\big)
= Q_i\big(q_i \mid \mathcal{H}_{i-1}\big)\, \hat{p}\big(r_i \mid q_i\big),
\end{equation}
whose right-hand side does not depend on $A$. Therefore,
\[
\Pr\big(q_i, r_i \mid A,\hat{p}, \mathcal{H}_{i-1}\big)
= \Pr\big(q_i, r_i \mid \hat{p}, \mathcal{H}_{i-1}\big)
\quad \text{for all } i.
\]
Plugging this equality back into the product expansion~\eqref{eq:chainrule_transcript} implies
\[
\Pr(T_k=t \mid A,c,\hat{p}) = \Pr(T_k=t \mid c,\hat{p}) \quad \text{for all } t,
\]
which proves~\eqref{eq:step2_target} and hence $A \perp T_k \mid (c,\hat{p})$, i.e., the Markov chain $A \to \hat{p} \to T_k$ (conditioning on $c$).

Finally, by the data processing inequality applied to $A \to \hat{p} \to T_k$ (conditioning on $c$),
\[
I(A;T_k\mid c) \le I(A;\hat{p}\mid c).
\]

\medskip
\noindent
\textbf{Consequence.} Combining Steps 1 and 2 yields both inequalities and also implies the conditional Markov chain $A \to \hat p \to T_k \to \hat R$ (conditioning on $c$).

\end{proof}

\subsection{Proof of Corollary~\ref{cor:fano}}
\label{app:proof_fano}

\begin{proof}
Condition on $c$ and let $m:=|R_c|$. Assume $A$ is uniform over
$R_c$, so $H(A\mid c)=\log m$. Let
\[
P_e := \Pr[\hat R\neq A\mid c],
\]
and define the error indicator
\[
E:=\mathbf{1}\{\hat R\neq A\}.
\]

Because $E$ is determined by $(A,\hat R)$,
\begin{align}
H(A\mid \hat R,c)
&= H(A,E\mid \hat R,c) \\
&= H(E\mid \hat R,c)+H(A\mid E,\hat R,c).
\end{align}
The first term satisfies
\[
H(E\mid \hat R,c)\le H(E\mid c)=h(P_e),
\]
where $h(\cdot)$ is the binary entropy. When $E=0$, we have
$A=\hat R$, so
\[
H(A\mid E=0,\hat R,c)=0.
\]
When $E=1$, the answer $A$ can take at most $m$ possible values,
including when $\hat R$ is an abstention or another output outside
$R_c$. Therefore,
\[
H(A\mid E=1,\hat R,c)\le \log m.
\]
It follows that
\[
H(A\mid \hat R,c)
\le h(P_e)+P_e\log m
\le \log 2+P_e\log m.
\]
Rearranging gives
\[
P_e
\ge
\frac{H(A\mid \hat R,c)-\log 2}{\log m}.
\]
Since $A$ is uniform over $R_c$,
\[
I(A;\hat R\mid c)
=H(A\mid c)-H(A\mid\hat R,c)
=\log m-H(A\mid\hat R,c).
\]
Hence,
\[
P_e
\ge
1-\frac{I(A;\hat R\mid c)+\log 2}{\log m}.
\]
Finally, Theorem~\ref{thm:dpi_tts} gives
\[
I(A;\hat R\mid c)\le I(A;\hat p\mid c),
\]
and therefore
\[
\Pr[\hat R\neq A\mid c]
\ge
1-\frac{I(A;\hat p\mid c)+\log 2}{\log m}.
\]
\end{proof}

\subsection{Answer Rate and Hallucination Rate}
\label{ap:conditional_error}

Corollary~\ref{cor:fano} bounds $\Pr[\hat R \neq A \mid c]$, which counts abstention as an error. The following corollary restates this result in terms of the answer rate and the overall hallucination rate, as used in Section~\ref{subsec:mi_bound}.

\begin{corollary}[Answer rate and hallucination rate]
\label{cor:conditional}

Assume $A$ is uniform over $R_c$ given $c$, and let $m := |R_c|$. For any compute-only TTS output $\hat R$ with abstention option $\bot \notin R_c$,

\begin{equation*}
\Pr[\hat R \neq A \wedge \hat R \neq \bot \mid c]
\;\ge\;
\Pr[\hat R \neq \bot \mid c] - \frac{I(A;\hat p \mid c) + \log 2}{\log m}.
\end{equation*}
\end{corollary}

\begin{proof}
Condition on $c$ throughout. Since $\bot \notin R_c$, every attempted answer is
either correct or incorrect, so
\begin{align}
\Pr[\hat R \neq A \wedge \hat R \neq \bot \mid c]
&= \Pr[\hat R \neq \bot \mid c] - \Pr[\hat R = A \mid c].
\label{eq:hallu_decomposition}
\end{align}
Corollary~\ref{cor:fano} states
$\Pr[\hat R \neq A \mid c] \ge 1 - \frac{I(A;\hat p \mid c) + \log 2}{\log m}$,
equivalently
\begin{equation}
\Pr[\hat R = A \mid c] \;\le\; \frac{I(A;\hat p \mid c) + \log 2}{\log m}.
\label{eq:accuracy_ceiling}
\end{equation}
Substituting~\eqref{eq:accuracy_ceiling} into~\eqref{eq:hallu_decomposition}
yields the claim.
\end{proof}

\begin{remark}

Corollary~\ref{cor:conditional} makes the tradeoff explicit. The subtracted term represents an information-dependent accuracy ceiling that compute-only scaling cannot increase (Theorem~\ref{thm:dpi_tts}). When this ceiling is small, a high answer rate necessarily implies a high lower bound on hallucinations. 
\end{remark}

\section{Use of LLMs}

Large language models (LLMs) were used during manuscript preparation only for minor language editing, such as correcting grammar and typos. They were not used for generating research ideas, designing the method, or producing original content.

\end{document}